\documentclass[10pt,twocolumn]{article}
\usepackage[margin=0.75in]{geometry}
\usepackage[utf8]{inputenc}
\usepackage{booktabs, graphicx, amsmath, amssymb, subcaption, caption}
\usepackage{circuitikz}
\usepackage{import}
\usepackage{array}
\usepackage{svg}
\usepackage{multirow}
\usepackage{threeparttable}
\usepackage{float}
\usepackage{balance}
\usepackage{tabularx}
\usepackage[numbers,sort&compress]{natbib}
\usepackage{enumitem}
\usepackage{authblk}
\usepackage[colorlinks=true, allcolors=blue]{hyperref}
\usepackage{caption}
\usepackage[table]{xcolor}

\title{How does downsampling affect needle electromyography signals? A generalisable workflow for understanding downsampling effects on high-frequency time series}

\author[1]{Mathieu J.L. Cherpitel}
\author[2,3]{Janne A.M. Luijten}
\author[1]{Thomas H.W. Bäck}
\author[3]{Camiel Verhamme}
\author[2]{Martijn R. Tannemaat}
\author[1]{Anna V. Kononova}

\affil[1]{Leiden Institute of Advanced Computer Science, Leiden, The Netherlands}
\affil[2]{Leiden University Medical Centre, Department of Neurology, Leiden, The Netherlands}
\affil[3]{Amsterdam University Medical Centre, Department of Neurology, Amsterdam, The Netherlands}
\date{}

\begin{document}

\twocolumn[
  \begin{@twocolumnfalse}
    \maketitle
    \begin{abstract}
        Automated analysis of needle electromyography (nEMG) signals is emerging as a tool to support the detection of neuromuscular diseases (NMDs), yet the signals' high and heterogeneous sampling rates pose substantial computational challenges for feature-based machine-learning models, particularly for near real-time analysis. Downsampling offers a potential solution, but its impact on diagnostic signal content and classification performance remains insufficiently understood. This study presents a workflow for systematically evaluating information loss caused by downsampling in high-frequency time series. The workflow combines shape-based distortion metrics with classification outcomes from available feature-based machine learning models and feature space analysis to quantify how different downsampling algorithms and factors affect both waveform integrity and predictive performance. We use a three-class NMD classification task to experimentally evaluate the workflow. We demonstrate how the workflow identifies downsampling configurations that preserve diagnostic information while substantially reducing computational load. Analysis of shape-based distortion metrics and classification performance degradation showed that, on the EMGLAB dataset, decimation with anti-aliasing filtering better preserves signal characteristics key to the classification of neuromuscular diseases compared to shape-aware downsampling strategies when using the \textit{tsfresh} feature set. The results provide practical guidance for selecting downsampling configurations that enable near real-time nEMG analysis and highlight a generalisable workflow that can be used to balance data reduction with model performance in other high-frequency time-series applications as well. 
    \end{abstract}
    \vspace{2em}
  \end{@twocolumnfalse}
]

\section{Introduction}
\label{sec:intro}
Needle electromyography (nEMG) is a diagnostic technique for neuromuscular disorders (NMDs). NMDs are diseases affecting muscles, nerves or motor neurons that cause weakness or sensory deficits~\cite{preston_electromyography_2020}. 
Timely and accurate diagnosis of NMDs is becoming more important as more treatments for various NMDs are rapidly emerging~\cite{chiapparoli_glance_2022, lehmann_chronic_2019, miller_trial_2022}.
During nEMG examination, a needle electrode is inserted into the muscle to evaluate its electrical activity at rest and during voluntary contraction. 
Interpretation of the nEMG is performed in real-time by a clinical neurophysiologist through audio-visual assessment of the signal~\cite{dumitru_electrodiagnostic_2001}. 
Even with adequate training and expertise, the interpretation of nEMG signals remains inherently subjective, varies across medical centres and can potentially result in limited agreement among different clinicians~\cite{narayanaswami_critically_2016}. 

To improve the objectivity and reproducibility of nEMG interpretation, quantitative nEMG approaches have been investigated for several decades \cite{stalbergOutliersWayDetect1994, stashukDecompositionQuantitativeAnalysis1999, sonooNewAttemptsQuantify2002, dohertyDecompositionbasedQuantitativeElectromyography2003}. These approaches aim to characterise motor unit action potentials and interference patterns using mathematically derived descriptors that support diagnostic decision-making. More recently, machine learning (ML) techniques have been increasingly developed to automate or assist nEMG interpretation ~\cite{de_jonge_artificial_2024}. Both conventional feature-based ML methods and deep learning approaches have demonstrated promising performance in distinguishing normal, neuropathic, and myopathic nEMG ~\cite{noderaDeepLearningWaveform2019, yoo_residual_2022, tannemaat_distinguishing_2023}. Feature-based ML approaches, in particular, offer superior explainability compared to deep learning approaches and have shown promising results in neurophysiological applications~\cite{de_jonge_artificial_2024, tannemaat_distinguishing_2023, hosseini_review_2021}.

The drawback of these feature-based ML approaches, however, is the need for extensive feature extraction before classification. 
Together with the high temporal resolution of nEMG signals, this poses a computational challenge and hinders (near) real-time analysis and, thus, clinical implementation. 

A possible strategy to mitigate this issue is to reduce the temporal resolution of the signal by using downsampling before applying feature extraction methods. 
However, downsampling introduces another trade-off: reducing the temporal resolution is only appropriate when the essential (diagnostic) information embedded in the signal is preserved. 

\begin{figure*}[t]
    \centering
    \includegraphics[width=0.9\linewidth]{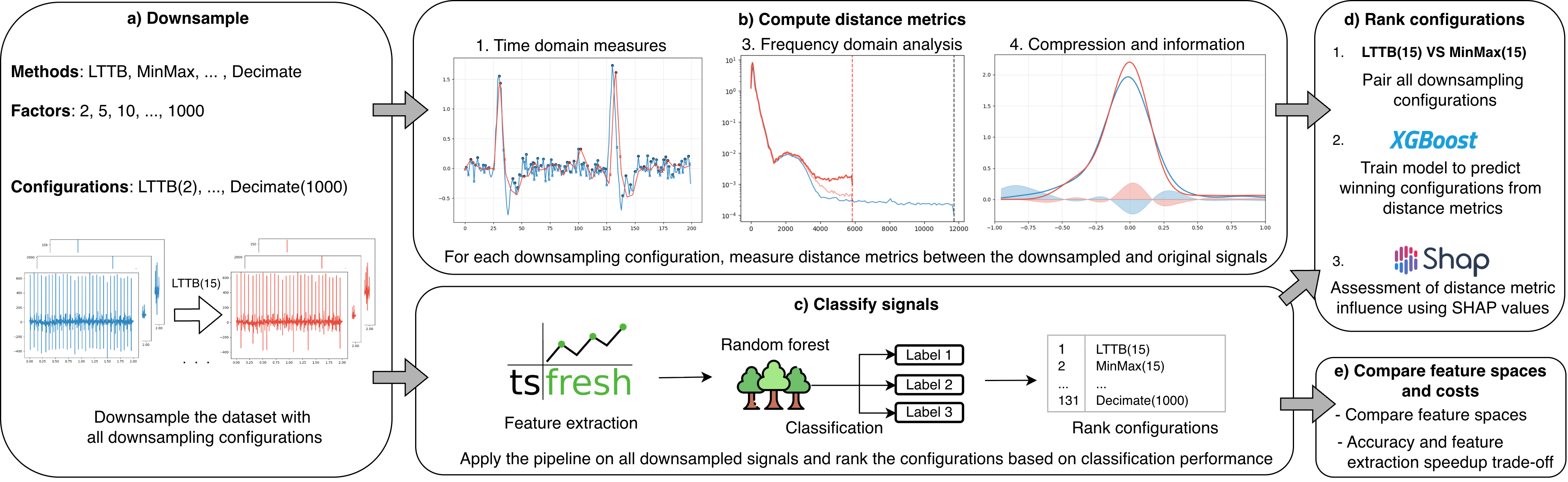}
    \caption{Five steps of the proposed workflow to investigate the effects of downsampling on time series.}
    \label{fig:workflow}
\end{figure*}

Previous studies within the biomedical field have explored downsampling in electroencephalography (EEG)~\cite{bischof_geometric_2021}, the electrocardiogram (ECG)~\cite{salimi_exploring_2025, singstad_assessing_2025}, and surface EMG~\cite{diab_effect_2013, ravier_refined_2021}. 
Collectively, these studies show that the impact of downsampling varies widely: it can reduce classification accuracy~\cite{bischof_geometric_2021}, while it may also preserve~\cite{salimi_exploring_2025} or even enhance~\cite{singstad_assessing_2025, diab_effect_2013} classification performance while reducing computational load. 
Using seven datasets from within and beyond the biomedical field, Moltó-Gallego et al. similarly reported that downsampling impact varies across extracted features, downsampling method, and analysed dataset~\cite{molto-gallego_enhancing_2025}. 
Additionally, they further demonstrated that downsampling can, in specific scenarios, simultaneously improve classification performance and reduce computational load. 

While downsampling is a widely recognised strategy to reduce computational load, its application across studies is non-standardised, and a systematic evaluation method to determine acceptable levels of information loss is lacking. This paper addresses this gap. It is organised as follows: Section \ref{sec:methods} presents a workflow that combines shape-based distortion metrics, classification performance, and feature space analysis to assess how downsampling affects high-resolution time series. In Section \ref{sec:exp_eval}, we apply this workflow to a three-class nEMG problem and present the specific results of that evaluation, including the effects on computational load. Section \ref{sec:discussion} discusses our results and reflects on the general applicability of the proposed workflow. Section \ref{sec:conclusion} concludes the paper and outlines the implications of our results for clinical implementation of nEMG analysis pipelines.

\section{Methods}
\label{sec:methods}
The proposed workflow to investigate the effects of downsampling on time series consists of five main steps, summarised in Figure~\ref{fig:workflow} and elaborated in this section.

\begin{enumerate}
    \item We first define several downsampling configurations, each combining an algorithm and a downsampling factor. Each configuration is then applied to all signals in the dataset, resulting in one downsampled dataset per considered configuration.
    \item We then compute a selection of distance metrics between each downsampled signal and its corresponding original signal for each configuration.
    \item In parallel, we train an automated time series classification pipeline on each dataset. This pipeline was originally designed to process vehicle crash data~\cite{koch_machine_2018, koch_machine_2018-1} and proved to offer good performance on a variety of clinically meaningful time series data, including EEG~\cite{koch_automated_2019} and nEMG~\cite{kefalas_automated_2020}. Once the pipeline is trained on each dataset separately, we evaluate the classification performances and rank the configurations according to the achieved accuracies.    
    \item Using the distance metrics and the classification performance based ranks, we train a ranking model to predict the best downsampling configurations from distance metrics alone. From this model, we extract SHAP values to interpret the contributions of individual distance metrics to identify key signal characteristics that the classifier relies on.
    \item Finally, we compare the feature spaces obtained from each dataset to analyse how they get altered across downsampling configurations. We report the reduction in feature extraction time and its trade-off with classification accuracy. 
\end{enumerate}

\subsection{Downsample}
\label{sec:downsample}
\begin{figure}[ht]
    \centering
    \includegraphics[width=0.9\linewidth]{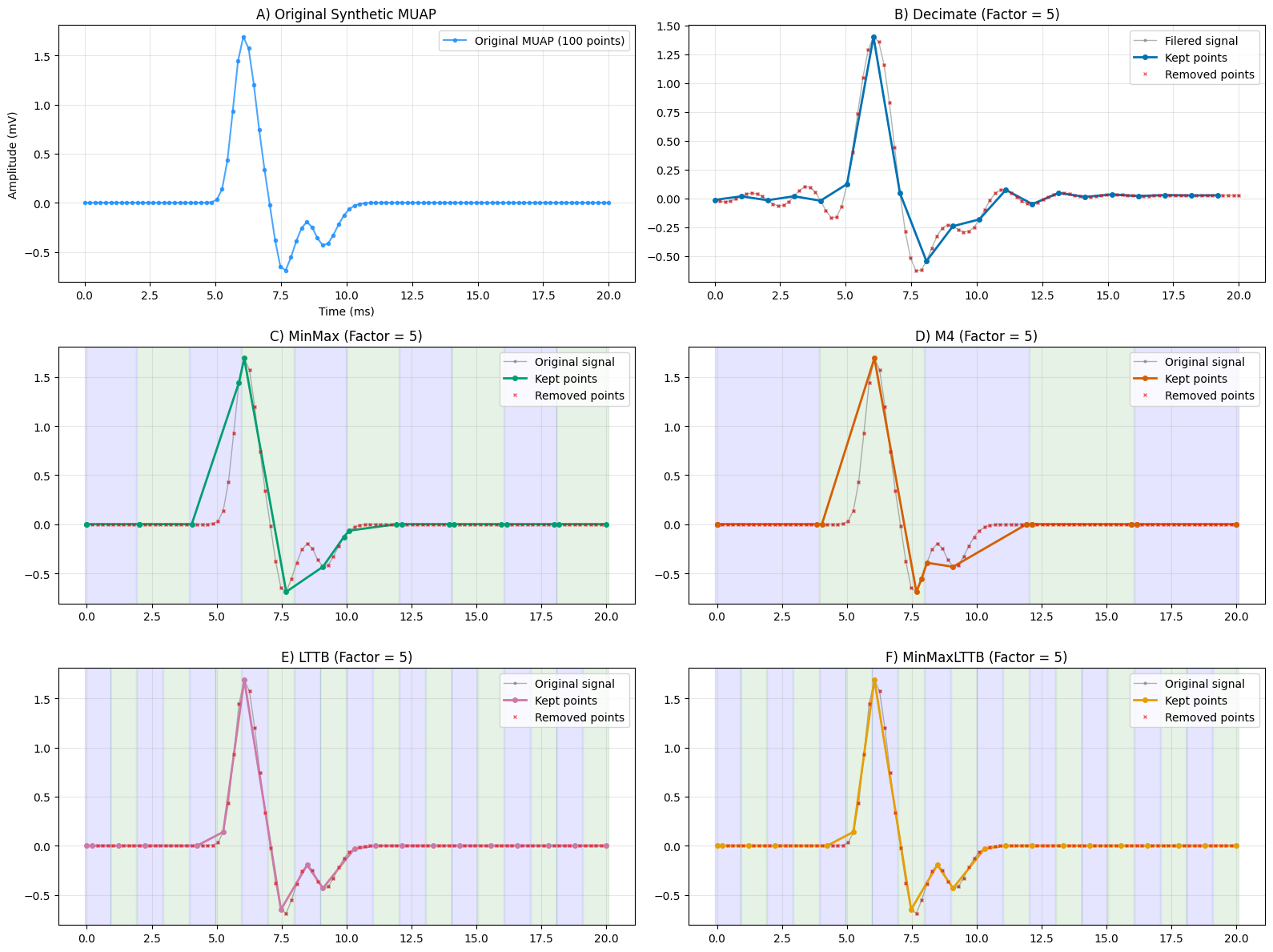}
    \caption{
    Effects of five downsampling methods on a synthetic MUAP-like waveform.
    Shaded regions indicate the value groups (when applicable) used by each algorithm.
    Typical effects are visible:
    \textbf{B)} Decimate preserves MUAP duration but loses peak amplitude and phase detail.
    \textbf{C)} MinMax preserves amplitude and duration but alters phase count.
    \textbf{D)} M4 maintains amplitude but distorts duration and phases.
    \textbf{E)} and \textbf{F)} LTTB and MinMaxLTTB yield the most faithful approximations, preserving amplitude, duration, and phase structure.
    }
    \label{fig:downsamplers}
\end{figure}

Downsampling reduces the temporal resolution of a time series through the selection of a subset of points. Given a time series $X = [x_1, x_2, \dots, x_N]$, the downsampled version is defined as:
\[
X' = d(X, k)
\]
where $d$ denotes the downsampling algorithm and $k$ the downsampling factor. The resulting signal $X'$ has a length $|X'| \approx \lfloor N/k \rfloor$, with both dimensionality and sampling frequency reduced by approximately a factor $k$. 

Although the most common approach to downsampling is a simple Decimate (retaining every $k$-th point), we also evaluate algorithms developed for data visualisation (MinMax, M4, LTTB and MinMaxLTTB). Since nEMG signals are traditionally audio-visually assessed, attempting to maintain their visual shape could preserve information content. These algorithms use groups of values from which a set number of points are kept based on their rules. In this case, the downsampling factor determines the size of the groups of values.

To illustrate the behaviour of the investigated algorithms, we generate a synthetic waveform resembling a motor unit action potential (MUAP) and apply each algorithm, yielding their respective downsampled versions. Figure~\ref{fig:downsamplers} shows the results for the five investigated algorithms:
\begin{itemize}
    \item \textbf{Decimate}: Standard decimation by retaining every $n$th point after applying an anti-aliasing filter.
    \item \textbf{MinMax}: This method retains extreme values by returning the $min$ and $max$ point from each group, aiming to preserve the overall amplitudes of the signal.
    \item \textbf{M4}: Proposed by Jugel et al.~\cite{jugel_m4_2014}, this algorithm aims to avoid pixel representation errors that can occur when using MinMax. By ignoring the transitions between groups, the line drawn between two extrema can include a pixel that was not used in the original representation of the data, resulting in a representation error. To overcome this problem, the algorithm selects both the boundaries ($first$, $last$) and the extreme values ($min$, $max$) of each group to ensure an accurate connection between groups. This approach requires groups twice the size of those for MinMax since the number of selected values increases from two to four. If the same point is selected multiple times (e.g., the last value is also the maximum), fewer than four points are kept.
    \item \textbf{LTTB}: The Largest Triangle Three Buckets method, proposed by Steinarsson~\cite{steinarsson_1979-_downsampling_2013}, quantifies point influence by maximising the \textit{effective area}, i.e. the area of the triangle formed by the evaluated point, the point selected in the previous group, and the median of the next group. This technique is widely adopted in data visualisation due to its accurate approximation of signal shape.
    \item \textbf{MinMaxLTTB}: Introduced by Van de Donckt et al.~\cite{donckt_minmaxlttb_2023}, this variant reduces the computational cost of LTTB by pre-selecting vertical extremes ($min$, $max$) before applying LTTB. The ratio controlling the number of points retained before LTTB can be adjusted. Following the author's suggestions, we use a pre-selection ratio of 4.
\end{itemize}

It is standard practice to apply an anti-aliasing filter before downsampling with the decimate algorithm to prevent high-frequency components from being misrepresented as lower-frequency components after downsampling. We used the \texttt{scipy.signal.decimate} function with a default FIR low-pass filter with zero phase. For the group methods, however, anti-aliasing filters remove signal features that these algorithms are designed to preserve: sharp peaks and vertical extrema. The filter will reduce the true maximum and increase the true minimum, yielding downsampled signals with underestimated peak-to-peak amplitudes. Since MinMax and M4 explicitly try to preserve these extremes ($min$, $max$), the filter corrupts the data they are intended to select. LTTB and MinMaxLTTB select points that maximise the effective area, which translates to prioritising points that define the sharpest turns or extremes in the signal shape. 
As these group methods do not sample uniformly, they inherently distort the signal's original frequency characteristics. Therefore, applying low-pass filtering before these algorithms provides no benefit and instead induces artificial smoothing.

Given a dataset $\mathcal{X}=\{X_1, X_2, ..., X_M\}$ of $M$ time series, a set of downsampling methods $\mathcal{D}$ and factors $\mathcal{K}$, each unique configuration $(d,k)$ with $d \in \mathcal{D}$ and $k \in \mathcal{K}$, is applied to all time series in the dataset. The complete collection of downsampled datasets is then:
\[
\{\mathcal{X}_{(d, k)} \mid d \in \mathcal{D},\ k \in \mathcal{K}\},
\]
Which, in addition to the original dataset, serves as input for subsequent distance metric analysis and classification.

\subsection{Compute Distance Metrics}
\label{sec:compute_distance_metrics}
To assess the impact of downsampling on the shape of our signals, we compute several distance metrics between each original signal and its corresponding downsampled version for all configurations. We selected metrics from the time domain, frequency domain, and information theory that capture key signal characteristics potentially affected by downsampling. The objective is to identify computationally efficient measures of information loss and signal distortion without requiring the full classification pipeline to be run for all configurations. The selected metrics are described in Table~\ref{tab:metrics}. They will be used to identify signal characteristics that are important for the classification model and which downsampling methods best preserve them.

\begin{table*}
\centering
\caption{Overview of the distance metrics used to compare original and downsampled signals.}
\begin{threeparttable}
\renewcommand{\arraystretch}{2.5}
\begin{tabular}{m{6.5cm} m{3cm} m{8cm}}
\toprule
\noalign{\vspace{-1.5ex}}
Distance metric & Type & Formula \\
\noalign{\vspace{-0.5ex}}
\midrule

Root Mean Squared Error (RMSE) & Time domain & 
\(\displaystyle RMSE(X,Y)=\sqrt{\frac{1}{N}\sum\limits_{i=1}^N(x_i-y_i)^2}  \hfill (1)\) \\[3ex]

Normalised Mean Squared Error (NMSE) & Time domain & 
\(\displaystyle NMSE(X,Y)= \frac{\sum\limits_{i=1}^N(x_i-y_i)^2}{\sum\limits_{i=1}^N(x_i-\overline{x})^2}  \hfill (2)\) \\[5.5ex]

Pearson Correlation Coefficient (PCC)$^*$ & Time domain & 
\(\displaystyle PCC(X,Y) = \frac{\sum\limits_{i=1}^n(x_i-\overline{x})(y_i-\overline{y})}
{\sqrt{\sum\limits_{i=1}^n(x_i-\overline{x})^2}\sqrt{\sum\limits_{i=1}^n(y_i-\overline{y})^2}}  \hfill (3)\) \\

Spearman Correlation Coefficient (SCC)$^*$ & Time domain & 
\(\displaystyle SCC(X,Y) = PCC(R(X),R(Y)) \hfill (4)\)\\

Pearson Envelope Correlation$^*$ & Time domain & \(\displaystyle PCC(\mathcal{H(X)}, \mathcal{H}(Y)) \text{ and } SCC(\mathcal{H(X)}, \mathcal{H}(Y)) \hfill (5) \) \\

Zero Crossing Rate (ZCR)$^{**}$ & Time domain & \(\displaystyle ZCR(X) = \frac{1}{|X|-1}\sum\limits_{t=1}^{|X|-1}|s[x(t)]-s[x(t-1)]| \hfill (6) \)\\

Peaks count$^{**}$ & Time domain & \(\displaystyle \#peaks(X) \text{ (scipy find\_peaks, default params)} \hfill (7)\) \\

Skewness$^{**}$ & Time domain & \(\displaystyle Skew(X)=\frac{\frac{1}{N}\sum\limits_{i=1}^N(x_i-\overline{x})^3}{(std(X))^3}  \hfill (8)\)  \\[3ex]

Kurtosis$^{**}$ & Time domain & \(\displaystyle Kurt(X)=\frac{\frac{1}{N}\sum\limits_{i=1}^N(x_i-\overline{x})^4}{(std(X))^4}  \hfill (9)\)  \\[2.5ex]

\midrule


Log Spectral Distance (LSD) & Freq. domain & \(\displaystyle \text{LSD}(X,Y) = \left( \frac{1}{N_f}\sum_{i=1}^{N_f} \left| \log \frac{P_X(f_i)}{P_Y(f_i)} \right|^p \right)^{\frac{1}{p}} \hfill (10)\) \\[3.5ex]

\midrule

Normalised Compression Distance (NCD) & Info theory & 
\(\displaystyle NCD(X,Y) = \frac{Z(XY) - \min(Z(X),Z(Y))}{\max(Z(X),Z(Y))} \hfill (11)\) \\

Jensen-Shannon Divergence (JSD) & Info theory & \(\displaystyle JSD(P||Q)=\frac{1}{2}D_{KL}(P||M)+\frac{1}{2}D_{KL}(Q||M) \hfill (12)\) \\

\bottomrule
\end{tabular}
\begin{tablenotes}

\item[] $^*$For consistency, correlation measures were inverted, so that lower values always indicate greater similarity, aligning them with the other distance-based metrics.

\item[] $^{**}$Distance computed as absolute difference: $d(X,Y) = |f(X) - f(Y)|$ where $f$ is the respective metric.

\item[] (4): $R(X)$ denotes the ranks of elements in $X$ sorted in ascending order.

\item[] (5): $\mathcal{H}$ denotes the Hilbert transform for envelope extraction before measuring the Pearson correlation.

\item[] (6): $s(x) = 1$ if $x \geq 0$ else $0$.

\item[] (7): With scipy v 1.14.1.

\item[] (10): Where $P_X(f_i)$ and $P_Y(f_i)$ are the values of the normalised Power Spectral Densities (PSDs) of signal $X$ and $Y$ at frequency bin $f_i$ estimated with the welch method. $N_f$ is the total number of bins.

\item[] (11): $Z$ approximates Kolmogorov complexity via gzip compression \cite{li_similarity_2004}.

\item[] (12): $D_{KL}$ = Kullback-Leibler divergence, $M = \frac{1}{2}(P+Q)$. PDFs estimated via histogram discretisation.

\end{tablenotes}
\end{threeparttable}
\label{tab:metrics}
\end{table*}

\subsection{Classify signals}
\label{sec:classify_signals}
The third step of our workflow evaluates how well downsampled signals support an automated analysis task. This step can be adapted to various tasks (e.g., classification, regression, or anomaly detection), depending on the research objective. In this study, we focus on a supervised classification task supported by an automated time-series classification pipeline~\cite{koch_machine_2018}. The pipeline consists of extracting features from each signal using the \textit{tsfresh} feature set~\cite{christ_time_2018} by Christ et al. The extracted features are then used to train and validate a Random Forest classifier using 5 repetitions of 3-fold stratified group cross-validation performed at the patient level. This splitting strategy ensures that all signals from the same patient are assigned exclusively to either the training or testing fold. Although a higher number of folds would have been preferable, the number of folds was limited by the small number of patients in the myopathic group to ensure that all classes were represented in both the training and validation splits.

During initial experiments, we observed high variance in classification performance across folds because some patients were considerably more difficult to classify. To ensure a fair comparison of the downsampling configurations, we fixed the random seeds to ensure that all configurations were evaluated on the same data splits. Classification performance was evaluated using balanced accuracy, per-class F1-score, per-class specificity and sensitivity, and one-vs-rest (OvR) ROC AUC. After evaluating the classification performance of each configuration on the validation splits, configurations were ranked according to their balanced accuracy scores.

\subsection{Rank configurations}
\label{sec:rank_conf}
To further understand the effects of downsampling, we train a ranking model using \textit{XGBoost}~\cite{noauthor_xgboost_nodate} to correlate distance metrics with classification performance. The objective is to identify which metrics best capture signal characteristics most relevant to the classifier, indicated by deterioration in classification accuracy.

Since training the complete pipeline to obtain the classification accuracy of each configuration is computationally expensive, we cannot generate enough samples to properly train a regression model directly predicting classification accuracy from the distance metrics. Instead, we adopt a pairwise ranking approach, which allows the model to learn relative performance between configurations from a limited number of evaluated samples. In this approach, all possible non-repetitive configuration pairs are generated within each unique factor group. This means that configurations sharing the same factor are compared against each other, i.e., downsampling algorithms are compared using the same downsampling factor. Each comparison is then used as a sample, consisting of the difference between the distance metrics of the two configurations and a binary label: 1 if the first configuration achieves a higher classification accuracy than the second (configuration 1 wins), and 0 otherwise (configuration 1 loses). In pairwise ranking, the achieved rank of a configuration is determined by the number of victories it obtains. Since each configuration is compared to all others using the same factor, this ensures an accurate intra-factor ranking. Once the \textit{XGBoost} model is trained, we extract SHAP values to evaluate the contributions of each distance metric.

\subsubsection{Evaluation of pairwise comparisons and consequent ranking}
\label{sec:eval_rank}
The pairwise ranking approach consists of creating non-repetitive pairs, predicting the winner of each pair, and ranking the configurations based on the number of victories within each factor. To evaluate the performance of a model trained for this purpose, two approaches can be used: (1) how often the model correctly identifies the winner (i.e., the configuration with the higher balanced accuracy), and (2) how accurate the final ranking is. For the latter, a correlation measure called \textit{Kendall’s $\tau$} assesses ordinal association between the true and predicted ranking at each factor. When evaluating how well the model predicts the correct winner, it is important to note that not all pairwise comparisons are equally difficult or relevant. Some pairs involve configurations with very different baseline performance, resulting in large differences in distance metrics and making correct classification trivial. To avoid overestimating the performance of the \textit{XGBoost} model due to these easy comparisons, we calculate the weighted \textit{XGBoost} accuracy based on an exponential decay of the difference in random forest classification accuracy. Pairs with smaller differences in random forest classification accuracy receive higher weights, increasing their impact on the weighted XGBoost accuracy.

For each pair $(i,j)$, let the difference in performance be:
\[
\Delta_k = |\text{acc}_i - \text{acc}_j| 
\]
A weight is assigned to each pair, using a decaying exponential function:
\[
w_k = e^{-\lambda\Delta_k}
\]
and the weighted accuracy of \textit{XGBoost} is measured as:
\[
\text{Weighted accuracy} = \frac{\sum_kw_k.C_k}{\sum_kw_k}
\]
Where $C_k$, the correctness, is equal to 1 if the model predicted correctly and 0 otherwise, and $\lambda$ is the decay rate of the weights.

\subsection{Compare feature spaces and costs}
\label{sec:feat_space_cost}
Since we use a feature-based classifier, this final step analyses how downsampling affects the high-dimensional feature spaces and computational cost associated with feature extraction. We use four techniques:
\begin{itemize}
    \item \textbf{Feature Importance Shifts}: Following classification, we obtain a vector for each feature, representing its importances across folds. To compare how feature importances change across downsampling configurations, the fold-averaged importances of the features are clustered using $k$-means clustering. The resulting clusters exhibit the global behaviour of feature groups and allow for interpreting shifts in feature relevance. 
    \item \textbf{Feature Importance Trajectories}: To visualise how feature importance evolves as the downsampling factor increases, the importance vectors for all configurations are projected to a lower-dimensional embedded space using dimensionality reduction techniques (e.g., MDS, t-SNE). This projection enables visualising trajectories for each downsampling algorithm, revealing how the feature relevance structure shifts as the downsampling factor increases.
    \item \textbf{Feature Extraction Speedup}: The primary benefit of downsampling is the reduction in feature extraction time. We measure feature extraction time $t_{ds}$ for each downsampling configuration. The speedup $S$ achieved is quantified by the ratio of the feature extraction time of the original signal ($t_{orig}$) to the downsampled signal: $S = t_{orig} / t_{ds}$. This provides a representative measure of the computational gain, enabling assessment of the trade-off with classification performance. 
\end{itemize}

This analysis focuses on feature importances because the large variability of feature values between samples complicates direct interpretation. 

\section{Experimental Evaluation}
\label{sec:exp_eval}
To illustrate and evaluate the workflow proposed in Section \ref{sec:methods}, we apply it to the EMGLAB dataset. The following subsections describe the dataset used, the experimental setup, and the obtained results.

\subsection{Data}
\label{sec:data}
To assess the impact of downsampling on nEMG, we use the EMGLAB dataset, as it provides high-quality, annotated contraction segments that facilitate the analysis of motor unit activity. The data was collected from the Department of Clinical Neurophysiology of Rigshospital, Copenhagen, for a study on signal decomposition and firing patterns analysis~\cite{mcgill_emglab_2005}. It has been shared publicly and is commonly used in a wide variety of studies due to its ease of use and the difficulty in obtaining other labelled nEMG signals. A scoping review by De Jonge et al.~\cite{de_jonge_artificial_2024} identified a total of 31 out of 51 papers that used EMGLAB as their primary dataset to develop artificial intelligence applications. Signals were recorded using a concentric needle inserted into the biceps brachii of 22 patients at a low voluntary level of contraction. All signals consist of 262124 measurements sampled at $23,437.5 \text{ Hz}$ (approximately 11 seconds). High- and low-pass filter boundaries were set to $2 \text{ Hz}$ and $10 \text{ kHz}$, respectively. The distribution of labels across patients is relatively balanced with 9 control, 6 myopathic and 7 amyotrophic lateral sclerosis (ALS) patients. However, the number of recordings per patient differs largely between the labels: each control subject has 30 recordings, whereas the number of recordings for myopathic and ALS patients varies between 2 and 25 per patient. As a result, the dataset has an imbalance factor of 2.76 for the control label (see Table \ref{tab:emglab-distribution}). The diagnostic labels were confirmed using both clinical and electrophysiological signs.

\begin{table}[t]
\centering
\caption{Distribution of labels in the EMGLAB dataset.}
\renewcommand{\arraystretch}{1.2}
\begin{tabular}{lrrr}
\toprule
Label & Control & Myopathic & ALS \\
\midrule
Number of patients & 9   & 6   & 7   \\
Number of recordings  & 270 & 107 & 98  \\
\bottomrule
\end{tabular}
\label{tab:emglab-distribution}
\end{table}

\subsection{Experimental setup}
\label{sec:exp_setup}
The five investigated downsampling algorithms ($\mathcal{D}$) introduced in Section \ref{sec:downsample} have been implemented in a Python library by Van de Donckt et al~\cite{van_der_donckt_tsdownsample_2025}, called \textit{tsdownsample}. These algorithms were selected as they each have unique characteristics that will result in diverse signal representations. To explore a wide range of possibilities while maintaining computational feasibility, we selected 26 downsampling factors ($\mathcal{K}$). The selected factors provide dense coverage of potentially optimal downsampling factors at the lower end of the spectrum (2–100), while also including higher values (200–1000) to evaluate the effects of extreme downsampling:
\[
\mathcal{D} = \{\text{Decimate, M4, MinMax, LTTB, MinMaxLTTB}\}
\]
\[
\mathcal{K} = \{2, 5, 10, 15, \dots, 95, 100, 200, 300, 400, 500, 1000\}
\]

\begin{table*}[t]
\centering
\caption{Classification performance reached at the maximum performance-preserving downsampling factor for each algorithm along with the Original, non-downsampled signals.}
\label{tab:best_downsampling_performance}
\begin{tabular}{cccccc}
\toprule
Algorithm & Factor & Balanced Accuracy & ALS F1-score & Control F1-score & Myopathic F1-score \\
\midrule
Decimate & 30 & 0.6718 $\pm$ 0.0766 & 0.6928 $\pm$ 0.0489 & 0.7850 $\pm$ 0.0597 & 0.5683 $\pm$ 0.1091 \\
LTTB & 25 & 0.6625 $\pm$ 0.0669 & 0.6508 $\pm$ 0.0487 & 0.7832 $\pm$ 0.0580 & 0.5809 $\pm$ 0.1199 \\
MinMax & 25 & 0.6683 $\pm$ 0.0681 & 0.6635 $\pm$ 0.0512 & 0.7804 $\pm$ 0.0584 & 0.5847 $\pm$ 0.1066 \\
M4 & 20 & 0.6712 $\pm$ 0.0701 & 0.6909 $\pm$ 0.0527 & 0.7832 $\pm$ 0.0579 & 0.5648 $\pm$ 0.1193 \\
MinMaxLTTB & 25 & 0.6661 $\pm$ 0.0575 & 0.6588 $\pm$ 0.0553 & 0.7826 $\pm$ 0.0560 & 0.5854 $\pm$ 0.0978 \\
Original & 1 & 0.6790 $\pm$ 0.0573 & 0.6741 $\pm$ 0.0611 & 0.7827 $\pm$ 0.0628 & 0.6057 $\pm$ 0.1009 \\
\bottomrule
\end{tabular}
\end{table*}

\subsection{Classification performance across configurations}
\label{sec:class_perf}

\begin{figure}
    \centering
    \includegraphics[width=0.9\linewidth]{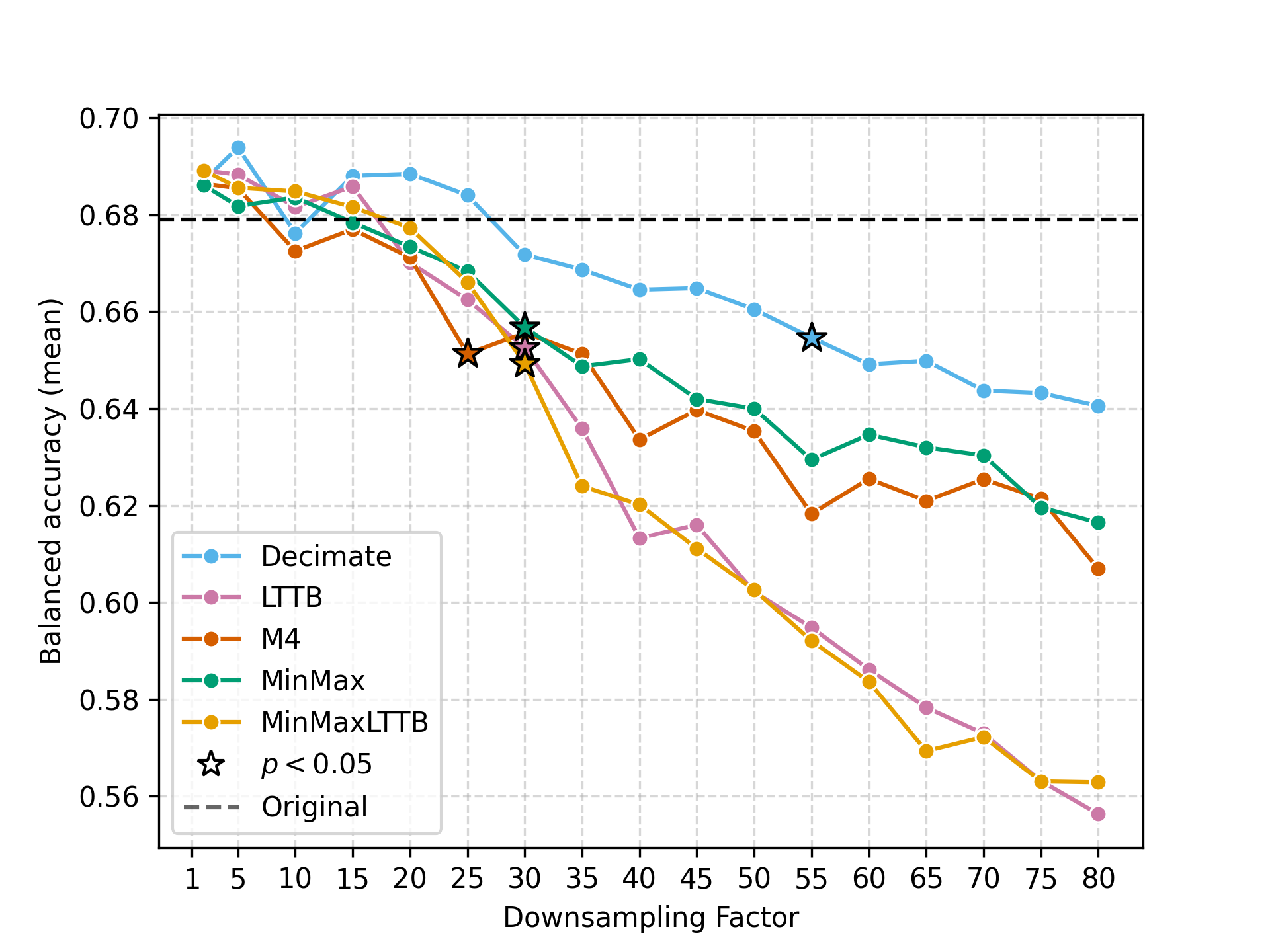}
    \caption{Achieved balanced accuracy across downsampling factors. Stars denote the first factor for which the balanced accuracy is significantly lower compared to the original (non-downsampled) signals.
    }
    \label{fig:balanced_acc}
\end{figure}

\begin{figure}
    \centering
    \includegraphics[width=0.9\linewidth]{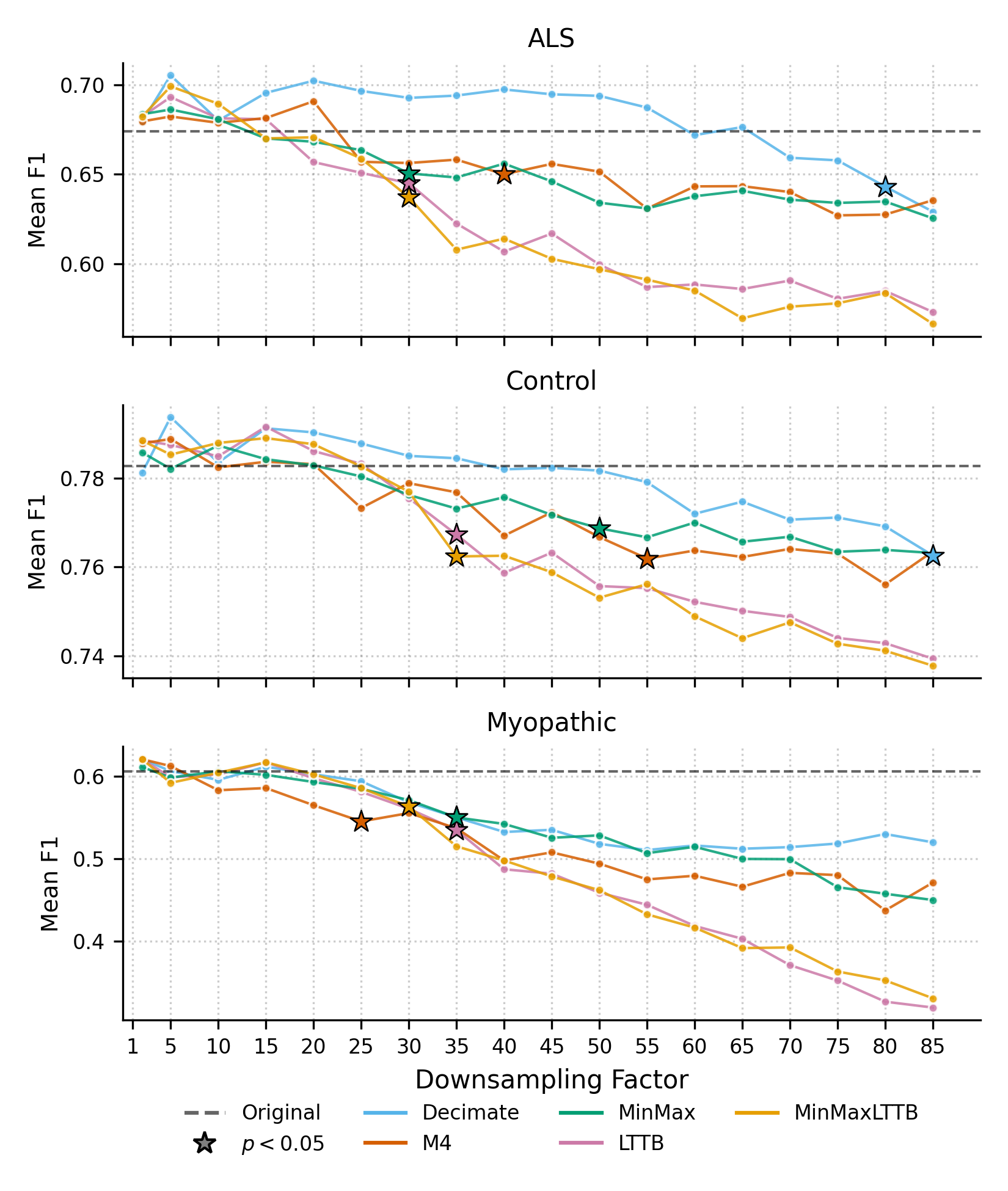}
    \caption{Class-wise mean F1-scores across downsampling factors. Stars denote the first factor for which the class-wise F1-score is significantly lower compared to the original (non-downsampled) signals.
    }
    \label{fig:f1_scores}
\end{figure}

As in previous work using the classification pipeline~\cite{kefalas_automated_2020}, EMGLAB recordings were segmented into two-second windows. They were subsequently downsampled and features were extracted using \textit{tsfresh} \cite{christ_time_2018}. A Random Forest classifier was trained on the resulting feature representations and performance metrics were collected as described in Section~\ref{sec:classify_signals}. Figure \ref{fig:balanced_acc} shows the evolution of balanced accuracy across downsampling factors for all downsampling algorithms, while Figure \ref{fig:f1_scores} presents the corresponding class-wise F1-scores. These metrics were chosen over traditional accuracy to address the class imbalance present at the signal level (Section \ref{sec:data}).

To evaluate statistical differences between each downsampling configuration and the original (non-downsampled) signals, a one-sided paired Wilcoxon signed-rank test was applied with the alternative hypothesis that performance was greater on the original signals. Pairing was performed across identical cross-validation splits (repetitions and folds) to ensure direct comparability. This test was applied separately for each downsampling configuration. To account for multiple comparisons across downsampling methods within each factor, Holm’s sequential correction procedure was applied to the resulting p-values. Correction was performed per factor, as each factor represents a distinct experimental condition. A global correction across all factors led to an overly conservative procedure due to the large number of comparisons across a very wide range of downsampling factors.

For each downsampling factor, a configuration was considered significantly degraded if any of the class-wise F1-scores (ALS, Control, or Myopathic) decreased significantly ($p < 0.05$ after correction) compared to the original signals. The corresponding \textit{critical downsampling factor} was defined as the smallest factor meeting this condition. M4 reaches its critical factor at 25 (triggered by Myopathic F1), while MinMax, LTTB, and MinMaxLTTB reach theirs at 30 (triggered by ALS F1, and Myopathic F1 for MinMaxLTTB). Decimate appears most resilient, reaching its critical factor at 35 (triggered by Myopathic F1). Table~\ref{tab:best_downsampling_performance} outlines performance at the maximum performance-preserving configuration (one step below the critical factor). The complete $p$-values measured and a critical difference plot at factor 35 are provided in ~\ref{appendix_a}.

\subsection{Ranking model results}
\label{sec:rank_model_res}
\begin{figure}
    \centering
    \includegraphics[width=1\linewidth]{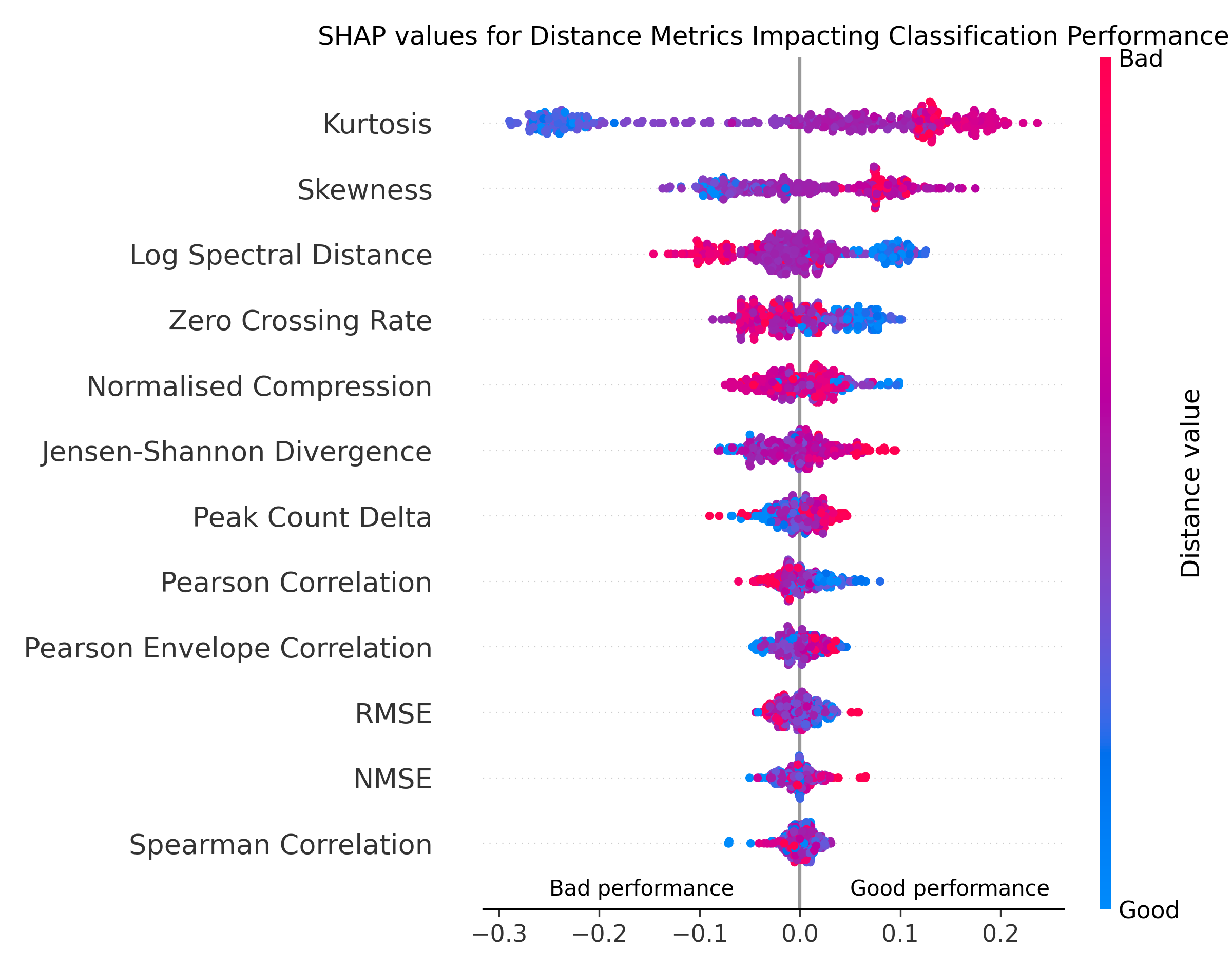}
    \caption{Reported SHAP values for the distance metrics used to predict winning downsampling configuration. For interpretability, the colour scale corresponds to the degree of similarity with the original signal: bad indicates a metric value representing greater dissimilarity from the original, while good indicates higher similarity. To ensure consistency across metrics, the values of correlation-based measures were inverted.}
    \label{fig:SHAP}
\end{figure}

The pairwise ranking model demonstrated viable predictive capabilities, particularly at higher downsampling factors. Specifically, we observe that the model begins to accurately rank downsampled configurations from factor 15 onwards (the full factor-by-factor ranking correlation breakdown is provided in ~\ref{appendix_b}). This was further supported by moderate weighted accuracies (0.754, 0.734, and 0.698 for decay rate of the weights: $\lambda=5, 10, 20$). While these scores reflect the inherent difficulty of the task, they demonstrate predictive utility even in challenging comparisons with small classification accuracy differences. Since the model's predictions rely on distance metrics, the SHAP analysis (Figure \ref{fig:SHAP}) reveals which measures of shape distortion are most informative for favourable downsampling.

\subsection{Effects of downsampling on the feature space}
As per the methodology detailed in Section 2.5, we analysed how downsampling affects the high-dimensional feature spaces and computational cost associated with feature extraction.

\subsubsection{Feature importance shifts}
\label{sec:imp_shifts}

\begin{figure}
    \centering
    \includegraphics[width=1\linewidth]{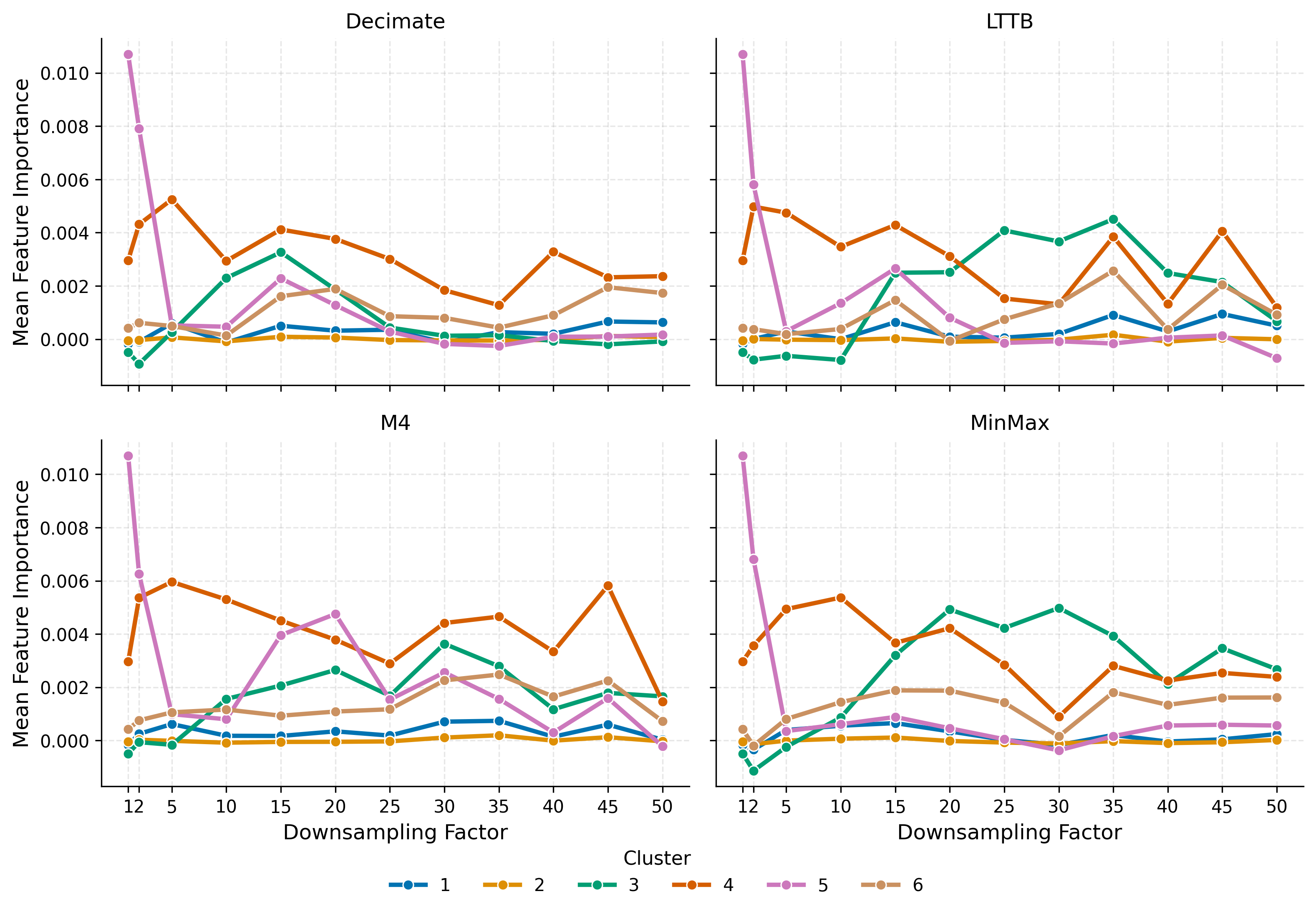}
    \caption{Feature importance clusters across increasing downsampling factors. The MinMaxLTTB algorithm is not present in this figure, as its  results were very similar to LTTB. For each algorithm, the first point represents the clustered feature importances of the original signals.}
    \label{fig:importance_shifts}
\end{figure}

To analyse how the importance of different features evolved across downsampling configurations, we clustered the fold-averaged feature permutation importance vectors using $k$-means clustering. Silhouette analysis revealed that six clusters best represent the evolution of feature importance for increasing downsampling factors; their trajectories are presented in Figure~\ref{fig:importance_shifts}.

Cluster 5 shows an apparent drop in importance across all algorithms. This cluster consists of a single feature ($\textit{number\_cwt\_peaks}$ with a width of five), which is estimated by smoothing the signal and counting peaks that are sustained over at least five samples above a Signal-to-Noise-Ratio (SNR) threshold. 
At the original sampling rate, MUAP peaks typically extend over five or more samples and are therefore detected by the feature. 
However, as seen in Figure \ref{fig:downsamplers}, these peaks are reduced to fewer than five samples after downsampling, preventing detection. 
This is further confirmed by the $\textit{number\_cwt\_peaks}$ feature with a width of one, which does not show a drop in feature importance after downsampling. 
Importantly, classification accuracy does not seem affected by Cluster 5 importance; classification accuracy remains stable during the drop seen for all algorithms

Cluster 4 showed the most robustness to downsampling in permutation importance averages. This cluster contains two features:
\begin{itemize}
    \item \texttt{agg\_linear\_trend(attr='stderr', chunk\_len=50, f\_agg='var')}
    \item \texttt{lempel\_ziv\_complexity(bins=100)}
\end{itemize}

This resilience likely reflects the fact that both features capture the broad, global shape of the signal rather than its small, high-frequency details. 

The \texttt{agg\_linear\_trend} feature measures how much the overall trend of the signal fluctuates across consecutive 50-sample windows. While downsampling tends to remove rapid, sharp variations, it largely preserves the global structural properties of the signal at moderate downsampling factors. Consequently, the overall shape and variation of the signal remain relatively stable.

Similarly, \texttt{lempel\_ziv\_complexity} evaluates the overall predictability and variety of patterns within the signal~\citep{lempelComplexityFiniteSequences1976}. By dividing the signal amplitude into 100 distinct levels (bins), it creates a detailed map of the signal's dynamic patterns. Although downsampling shortens the total signal length, it preserves the relative proportions of these patterns and the overall complexity of the physiological source. 

Despite intensive downsampling, these features remained useful indicators for accurate classification of nEMG signals.

\subsubsection{Feature importance trajectories in embedded space}
\label{sec:feat_imp_embed}
To understand how downsampling alters what the model relies on to make predictions, we analysed the evolution of feature importances. In this context, feature importance represents the contribution of the signals' features to the classification decision. By tracking how these importances shift, we can determine whether a downsampling algorithm preserves the underlying importances or forces the model to rely on different features as the downsampling factor increases.

Each downsampling configuration produces a high dimensional feature importance vector for every cross validation fold. To visualise the importance shifts, we embedded these vectors into a low-dimensional space. We then connected the points representing the same fold and algorithm in order of increasing downsampling factor. This creates trajectories that we can visually compare to compare the different downsampling algorithms and the original importances.

Given the substantial reduction from hundreds of selected features to 3 dimensions, selecting an embedding that truthfully represents the data was crucial. We evaluated several dimensionality reduction methods (PCA, KernelPCA ~\cite{scholkopf_kernel_nodate}, UMAP ~\cite{mcinnes_umap_2020}, Isomap ~\cite{tenenbaum_global_2000}, MSD ~\cite{kruskal_nonmetric_1964}, Spectral ~\cite{zelnik-manor_self-tuning_2004}, t-SNE ~\cite{van_der_maaten_viualizing_2008}) based on their ability to preserve global pairwise distances, measured by Pearson's and Spearman's correlations. Multidimensional Scaling (MDS) was selected as it achieved the highest fidelity, ensuring that the relative distances in the high-dimensional space are preserved.

\begin{figure}
    \centering
    \includegraphics[width=1\linewidth]{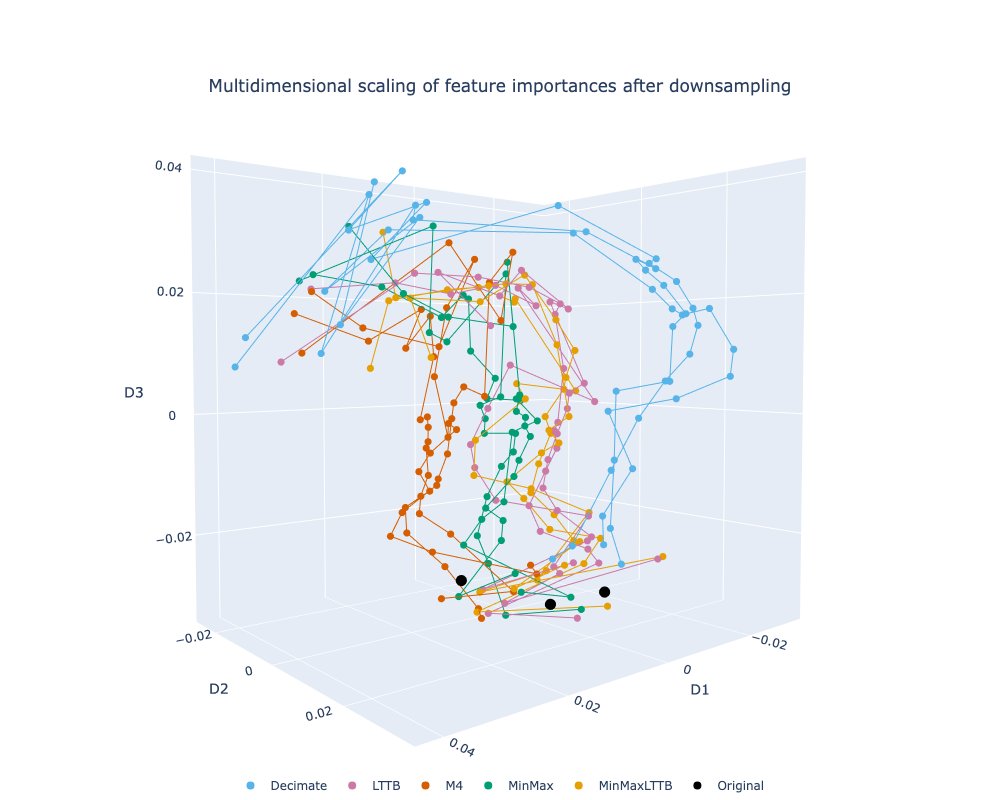}
    \caption{Trajectories of feature importances in the 3D embedded feature space using MDS, based on the results of downsampling on the EMGLAB dataset. Each point represents the feature importance vector for a specific downsampling configuration and cross-validation fold. Points are connected to form trajectories, illustrating the evolution of feature importances as the downsampling factor increases for each method.}
    \label{fig:embed}
\end{figure}

As seen in Figure \ref{fig:embed}, up to a downsampling factor of 5, the feature importance trajectories for all algorithms except for Decimate are similar, consistent with their comparable classification performance at low factors. Beyond factor 5, the trajectory of Decimate diverges from the shape aware downsampling algorithms, marking a turning point where the algorithm's downsampling rules substantially impact the resulting features. Decimate follows a distinctly different trajectory from the shape-aware algorithms. Finally, for all algorithms, increasing the downsampling factor causes a progressive increase in the variance of feature importances between folds, indicating a less consistent feature relevance pattern.

\subsubsection{Feature extraction speedup}
\label{sec:extr_speedup}

The classification pipeline uses \textit{tsfresh} for comprehensive time-series feature extraction. This strategy enables automated analysis while maintaining a high degree of explainability, as the model’s feature importances can be interpreted. 
However, this approach introduces significant computational load, since the initial feature extraction step is typically time-intensive. 
As the initial features are computed directly from the signal samples, reducing the number of data points greatly decreases the computational load. 
Since downsampling is an approximation of the original signal, we have seen in the classification performance results that after a certain factor, it can negatively impact classification performance. 

\begin{figure}
    \centering
    \includegraphics[width=1\linewidth]{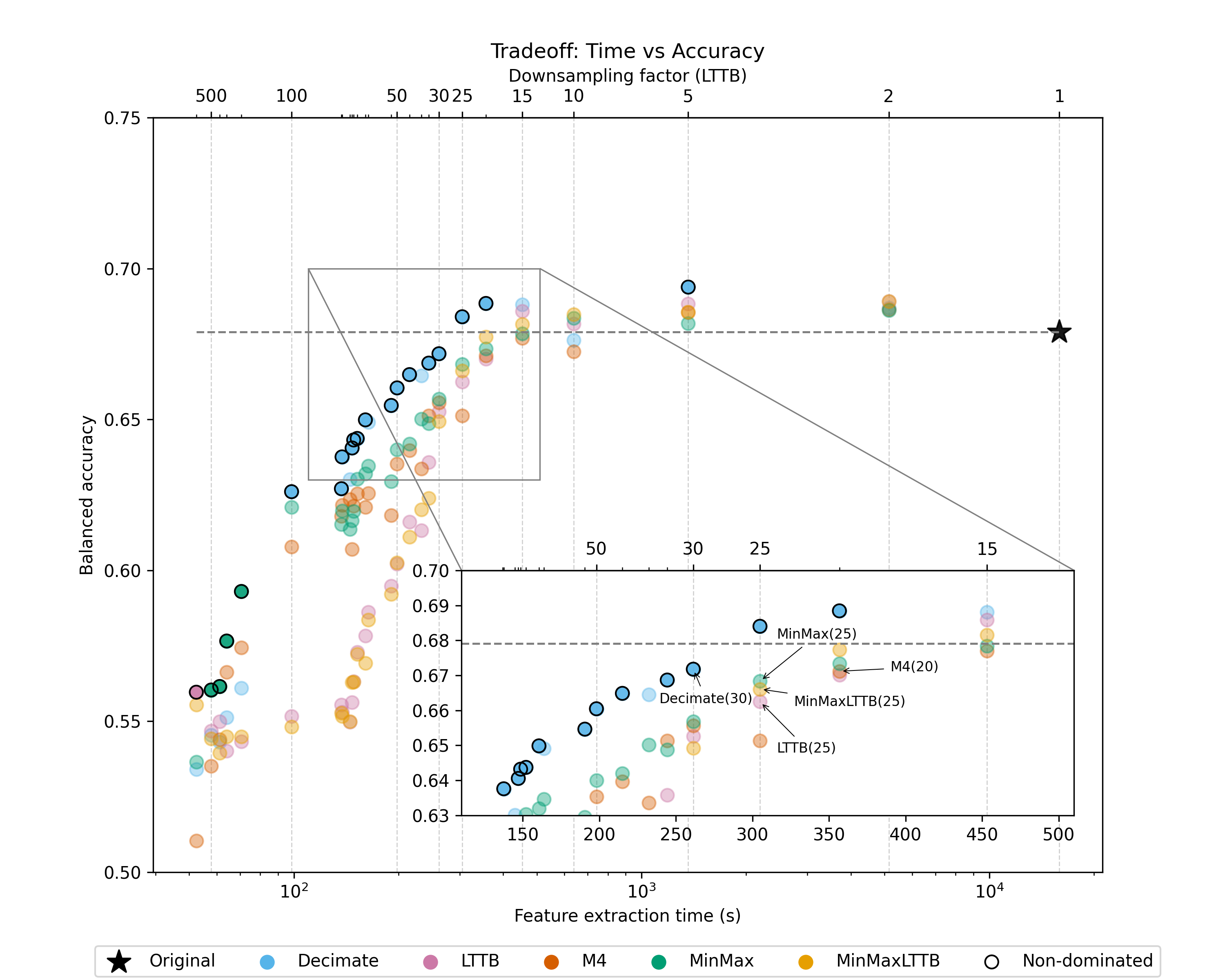} 
    \caption{Trade-off between measured feature extraction time and classification accuracy for each downsampling setup. Log scale is used for the time axis to better visualise the wide range of extraction times. Highlighted points are non-dominated solutions. The top axis denotes the LTTB downsampling factor; extraction times are comparable across methods at identical factors due to similar output cardinalities. The annotated points correspond to the critical factors for each method. Notably, the critical factors of Decimate appear as non-dominated solutions, signifying that despite their critical loss in performance, the achieved speedup represents a good trade-off.}
    \label{fig:tradeoff}
\end{figure}

To find the best downsampling configurations, we measure the feature extraction time for each downsampling configuration and visualise the trade-off between computation time and classification accuracy using a Pareto plot, presented in Figure~\ref{fig:tradeoff}. 
This figure highlights the non-dominated solutions, identifying the best trade-off between computation time and balanced accuracy. These solutions are the points where the extraction speed cannot be further increased without a corresponding loss in classification performance.
We mark the critical downsampling factors as explained in Section \ref{sec:class_perf} for each method. 
We observed an almost exponential decrease in feature extraction time up to a downsampling factor of 50, beyond which further downsampling yielded minimal computational gains.
The original signal is notably absent from the non-dominated solutions. This confirms that even when balanced accuracy is the primary goal, downsampling provides a more efficient representation of the data without sacrificing classification performance.
Between factors 2 and 100, virtually all non-dominated solutions correspond to the Decimate method, highlighting the superiority of this approach in our experiment. 
At factor 100 and above, the non-dominated solutions were produced by MinMax and LTTB, indicating that under extreme downsampling, these methods better preserve the information content relevant to our classification problem.
Table \ref{tab:tradeoff} numerically summarises the mean classification metrics and feature extraction speedup for each method, using the highest downsampling factor that maintains accuracy similar to the Original signal (i.e., right below the \textit{critical downsampling factor}).

\begin{table*}
    \centering
    \caption{Summary of classification performance (mean $\pm$ standard deviation) and feature extraction speedup average (expressed as x times faster) for downsampling algorithms. Speedup variance between algorithms with the same factor is influenced by cluster computation speed. Metrics are derived from 3-fold cross-validation with 5 repetitions using different seeds on the EMGLAB dataset. The reported downsampling factor for each method represents the maximum value achieved before a statistically significant critical difference in F1-score ($p<0.05$) for any of the three classes compared to the Original (1) signal.}
    \begin{tabular}{lcccccc}
        \toprule
        Algorithm (Factor) & Speedup(x) & Accuracy & F1 Score & Precision & Recall & ROC AUC \\
        \midrule
        \textbf{Decimate (30)} & \textbf{60.9} & $0.672 \pm 0.077$ & $0.721 \pm 0.068$ & $0.752 \pm 0.027$ & $0.672 \pm 0.077$ & $0.886 \pm 0.031$ \\
        LTTB (25) & 52.2 & $0.662 \pm 0.067$ & $0.715 \pm 0.065$ & $0.735 \pm 0.044$ & $0.662 \pm 0.067$ & $0.882 \pm 0.033$ \\
        MinMax (25) & 52.2 & $0.668 \pm 0.068$ & $0.718 \pm 0.063$ & $0.736 \pm 0.050$ & $0.668 \pm 0.068$ & $0.881 \pm 0.029$ \\
        M4 (20) & 44.6 & $0.671 \pm 0.070$ & $0.720 \pm 0.066$ & $0.742 \pm 0.047$ & $0.671 \pm 0.070$ & $0.884 \pm 0.033$ \\
        MinMaxLTTB (25) & 52.2 & $0.666 \pm 0.057$ & $0.717 \pm 0.060$ & $0.736 \pm 0.046$ & $0.666 \pm 0.057$ & $0.881 \pm 0.033$ \\
        Original (1) & 1.0 & $0.679 \pm 0.057$ & $0.728 \pm 0.058$ & $0.733 \pm 0.060$ & $0.679 \pm 0.057$ & $0.889 \pm 0.033$ \\
        \bottomrule
    \end{tabular}
    \label{tab:tradeoff}
\end{table*}

\section{Discussion}
\label{sec:discussion}
We developed a workflow to assess how downsampling affects the shape and classification performance of time series data. 
Applying this workflow to a popular dataset used for classifying EMG signals revealed that specific downsampling configurations can yield substantial feature extraction time speedup while maintaining classification performance. 
For a general recommendation, we suggest selecting the algorithm and factor combination that achieves the maximum speedup while maintaining classification accuracy. This corresponds to the highest downsampling factor before the accuracy exhibits a critical decline. When more specialised needs are present (i.e. emphasising speedup or accuracy), the solutions highlighted with a dark border in Figure \ref{fig:tradeoff} present the best found trade-offs.

Our experimental evaluation showed that the Decimate algorithms preserved classification accuracy up to and including factor 30, resulting in a feature extraction speedup of approximately 60-fold. Although we consider this configuration ideal, downsampling by a factor of 5 with any algorithm already resulted in a tenfold speedup without adversely affecting performance. 
The primary objective of this study is to understand the effects of downsampling, but solutions achieving slightly better speedup/accuracy trade-offs may be obtained by increasing the granularity of the downsampling factors.

Notably, the F1-score analysis presented in Figure \ref{fig:f1_scores} shows that the myopathic class was the most affected by downsampling, making it the limiting factor in determining the maximum performance-preserving downsampling factor. The decimation algorithm preserved F1-scores within an acceptable range for the ALS and control classes, even at relatively high downsampling factors (80-85). However, the myopathic class restricted the downsampling to factor 30 for Decimate and was the limiting factor for all downsampling strategies. This observation is clinically plausible, since myopathic MUAPs are often of short duration and severely polyphasic, which is a signal characteristic that requires a sufficiently high sampling frequency to be observed. 

Additionally, the trained ranking model can be used to quickly identify more precise factors that could produce improved results without having to run the entire expensive pipeline. The most important finding of our analysis is the contrast between the amount of downsampling permitted by each algorithm before damaging the information content of the signals. While LTTB and MinMaxLTTB allowed for substantial downsampling, M4 and the commonly used decimation algorithm caused a statistically significant decline in classification accuracy from factor 10 onwards.

These findings are consistent with previous literature reporting heterogeneous effects of downsampling time series on classification accuracy and emphasise the importance of selecting an appropriate downsampling configuration~\cite{bischof_geometric_2021, salimi_exploring_2025, singstad_assessing_2025, diab_effect_2013, molto-gallego_enhancing_2025}. 

To gain insight into these heterogeneous effects, our workflow extends existing work by introducing an analysis of the signal shape before and after downsampling. We quantified changes in signal shape by computing distance metrics before and after downsampling and evaluated their impact on classification performance using a ranking model. 

\subsection{SHAP analysis}
The ranking model achieved relatively high weighted accuracies and was able to accurately rank downsampling algorithms in classification performance from factor 15 onwards (Figure \ref{fig:ranking_corr}), indicating that the distance metrics employed can capture loss in classification accuracy.
The SHAP analysis (Figure \ref{fig:SHAP}) of this model suggests that preserving frequency domain and temporal-rate characteristics (specifically via the LSD and ZCR) is essential for accurate classification of the EMGLAB dataset. Conversely, strict preservation of the amplitude distribution (via kurtosis and skewness) shows an inverse relevance and is generally associated with lower classification performance. This indicates that preserving statistical outliers and distribution shapes via shape-aware algorithms may not retain the primary signal characteristics necessary for neuromuscular disorder diagnosis.

This methodology of altering the underlying signal properties to observe changes in ranking performance offers a promising framework for interpreting machine learning model behaviour. This framework diverges from traditional feature importance metrics by actively searching for characteristics of the data that affect predictive performance. Conceptually, similarly to an adversarial attack, by distorting the data and quantifying classification performance degradation, we gain deeper behavioural insights into the signal properties the classifier relies on. 

Our SHAP analysis showed that classification performance primarily relies on preserving the overall frequency content, as captured by spectral density and zero crossing rate, rather than on preserving extreme amplitudes. Crucially, these behavioural properties represent model characteristics that are not captured by conventional analysis of static \textit{tsfresh} feature importances.

\begin{table*}[t]
    \centering
    \small
    \caption{Summary of results from the investigated downsampling algorithms on EMGLAB}
    \renewcommand{\arraystretch}{1.2}
    \begin{tabularx}{\textwidth}{
        l
        >{\raggedright\arraybackslash}X
        >{\raggedright\arraybackslash}X
        c
        >{\raggedright\arraybackslash}X
        >{\raggedright\arraybackslash}X
    }
        \toprule
        Algorithm & Mechanism & Signal Impact & Critical Factor* & Main Advantage & Main Limitation \\
        \midrule
        Decimate & Uniform subsampling with anti-aliasing filter & Preserves low-frequency components & 30 & Highest accuracy retention, up to 60 $\times$ speedup & Loss of shape amplitude \\
        \hline
        LTTB / MinMaxLTTB & Area-based triangle bucket selection & Preserves envelope, peaks, and amplitude distribution & 25 & Best preservation of the visual shape of the signals & Distorts frequency content \\
        \hline
        MinMax & Two-point extrema selection (min, max) & Preserves large amplitudes and overall morphology & 25 & Low computational complexity & Limited frequency fidelity \\
        \hline
        M4 & Four-point extrema selection (min, max, first, last) & Strongly alters MUAP shape & 20 & Accurate transition between groups & Significant classification accuracy degradation \\
        \bottomrule
    \end{tabularx}
    \begin{tablenotes}
        \item [] * First downsampling factor resulting in a critical difference with the original signals in F1-score for one of the three classes.
    \end{tablenotes}
    \label{tab:downsampling_comparison}
\end{table*}

\subsection{Frequency analysis}
The observation that downsampling the EMGLAB signals using the standard decimation algorithm (with an anti-aliasing filter) up to factor 30 does not significantly degrade the performance of the machine learning pipeline (\textit{tsfresh} + Random Forest) raises important questions regarding the informative frequency range required for accurate classification of normal, myopathic and neuropathic nEMG signals. The original EMGLAB signals are sampled at $f_s = 23,437.5\text{ Hz}$ and band-pass filtered with cut-off frequencies of $2\text{ Hz}$ and $10\text{ kHz}$. Downsampling by a factor of 30 reduces the effective sampling rate to $781.25\text{ Hz}$, limiting the observable frequency range to $390.625\text{ Hz}$, a new Nyquist frequency. 

Because the achieved classification performance remains stable under this restricted bandwidth, the highest frequency components appear to be of limited relevance within the context of our current pipeline, supporting the hypothesis that these signals may be oversampled for standard automated classification. However, this claim must be nuanced by the fact that the pipeline achieves a baseline of approximately 70\% balanced accuracy. Consequently, it remains possible that informative features exist in the higher frequency bands, but these are either not adequately captured by the \textit{tsfresh} feature set or underutilised by the Random Forest classifier. Overall, rather than implying that higher frequencies are irrelevant, these results suggest that the features driving the model's diagnostic performance are concentrated within a lower frequency band.

\subsection{Feature space analysis}
Our visualisation of feature importance in the embedded space reveals a critical difference in the selection strategy across the different downsampling algorithms. Starting from factor 5, Decimate follows a distinctly different path compared to the shape-aware methods (LTTB, MinMaxLTTB, MinMax, M4), suggesting that its sampling procedure modifies the feature importance space in a unique way. At higher downsampling factors, the trajectories exhibit larger deviations between folds, indicating increased sensitivity of the extracted feature importance to both the sampling method and the specific data partition.
The close alignment of the LTTB and MinMaxLTTB feature space trajectories confirms that the MinMax filter successfully reduces the complexity of the LTTB selection procedure without introducing feature distortion. Although the computational cost associated with the downsampling is minimal in our classification pipeline, these results validate the use of that methodology.

Under extreme downsampling, feature importance variance between cross-validation folds notably increased. This indicates that as the data degrades severely, the model's reliance on features becomes unstable, leading to inconsistent predictions. Since most of the computational load of the model pipeline is the feature extraction time, we evaluated the gain obtained at each factor to identify the optimal time-accuracy trade-off. We observed an almost exponential decrease in feature extraction time until a factor of 50, after which the computational gain from further reduction became minimal. Crucially, using the best identified non-dominated downsampling configuration of decimation with an anti-aliasing filter at factor 30 yielded a significant speedup of approximately 60 times with no statistically significant loss in classification performance, strongly demonstrating the potential of appropriate downsampling strategies to reduce the computational load of automated feature-based classification. A summary of the findings made for each downsampling algorithm is presented in Table \ref{tab:downsampling_comparison}.

\subsection{Limitations}
An important limitation of this study is the use of a single dataset, as downsampling effects may vary between datasets~\cite{molto-gallego_enhancing_2025}. 
Moreover, since this dataset used a single sampling frequency, our results on the optimal downsampling factor are specific to this dataset and cannot be generalised to nEMG as a modality. More broadly, while the proposed workflow may be transferable as an evaluation procedure to understand downsampling effects, its outputs such as the preferred downsampling method, downsampling factor, and associated trade-off between computational cost and classification performance are potentially highly dataset specific.

The dataset itself introduces additional limitations: it contains only moderate voluntary contraction data recorded under controlled conditions, and its class distribution does not represent the typical clinical population. 
Specifically, these signals do not include rest or maximum voluntary contraction segments, or real-world clinical noise such as movement artefacts or electrode instability. 
The absence of rest segments is particularly unfavourable in our study, since these may contain higher frequency content such as abnormal spontaneous muscle fibre activity. These segments are therefore more likely to be affected by downsampling.

In addition, our conclusions are limited to a single feature representation and a single downstream machine-learning task. Different feature sets, modelling approaches, or other clinically motivated analysis tasks may respond differently to downsampling and could lead to different recommendations regarding the most appropriate downsampling strategy.

Finally, our workflow does not include comprehensive evaluation by clinical neurophysiologists. Although preservation of balanced classification performance and signal-shape characteristics provides evidence that diagnostically relevant information may be retained after downsampling, the resulting signals must ultimately remain interpretable for clinicians. Validation on independent nEMG datasets, ideally covering different sampling frequencies, contraction levels, and recording conditions, as well as expert clinical assessment of downsampled signals, is therefore needed before broader recommendations can be made on the range of informative frequency content in nEMG. However, such independent nEMG datasets are currently rare, highlighting a key bottleneck for a more complete validation.

\subsection{Future work}
The developed workflow combines downsampling neurophysiological time series with a three-class classification task. 
Beyond downsampling, the workflow can be adjusted to systematically evaluate other methods that change signal morphology, such as filtering or compression, and can be paired with any classification task. 
The workflow additionally enables the development of adaptive downsampling methods for specific signal types within and beyond the biomedical domain. 
Moreover, the development of a DICOM standard ~\cite{halford_standardization_2021, battaglia_neurophysiology_2024} for neurophysiological data further emphasises the relevance of our workflow, as it may contribute to the standardisation and interoperability of neurophysiological data. 
Because the workflow incorporates compression-related distance metrics, adapting it to analyse compression-induced signal distortions is a natural next step. 

Within our project, ARISE-NMD, we are focused on automated real-time analysis of nEMG signals and optimising our downsampling configuration is the first step. 
To address limitations regarding the dataset and clinical applicability, the workflow should be applied to a real-world nEMG dataset that includes rest segments, as these are critical to capture downsampling effects on high-frequency components. 
Ideally, multiple datasets with varying sampling frequencies and balanced class distributions should be used to evaluate the generalisability and robustness of the workflow. 
Additionally, clinical neurophysiologists who routinely evaluate nEMG signals should be involved to determine which morphological changes induced by downsampling remain acceptable for diagnostic purposes.

\section{Conclusion}
\label{sec:conclusion}
We presented a workflow to assess how downsampling affects automated analysis of nEMG signals. By analysing signal-shape distortion, classification performance, and computational cost, the workflow provides a standardised way to evaluate how different downsampling configurations influence both data quality and downstream time-series analysis. The workflow fills an important gap in the literature by introducing a systematic approach to understand the effects of downsampling on signal morphology and highlights which signal properties are most sensitive to the applied transformation in a reproducible and interpretable manner.

Beyond methodological value, our findings have direct implications for the real-time analysis of nEMG signals in clinical settings. 
The presented workflow offers a systematic way to identify downsampling strategies that achieve this balance, ensuring that feature-based models can run in near real-time without sacrificing predictive performance. 
As such, our study provides a foundation for developing clinically deployable nEMG analysis pipelines that meet the practical requirements of real-life conditions.

\section{Code and data availability}
\subsection{Code availability}
The pipeline code and scripts used to create the figures are available on GitHub at \url{https://github.com/ARISE-NMD/EMGLAB_downsample}. The specific version used to produce the results in this paper is archived at \url{https://doi.org/10.5281/zenodo.21028354}.
\subsection{Data availability}
The data generated by the pipeline during this study are available here: \url{https://doi.org/10.5281/zenodo.21028294}.

\section{Funding}
This work was supported by The Dutch Research Council (NWO) under the Open Technology Programme with file number 20852. 

\section{Acknowledgments}
The authors gratefully acknowledge the support of the following individuals and institutions: Michal Holub and Robert Bell (Cadwell Industries) for sharing their valuable knowledge on the application of downsampling in clinical devices, Prof. dr. Alfred C. Schouten (Delft University of Technology) for his expertise on sampling effects, Jonathan J. Halford (Ralph H. Johnson Department of Veterans Affairs Medical Center) for his comments on an early manuscript, and Jeroen J. Briaire (Leiden University Medical Center) for his advice on spectral analysis.

\balance
\bibliographystyle{unsrt}
\bibliography{references}

\appendix
\onecolumn

\section{Detailed statistical results for downsampling effects on classification performance.}
\label{appendix_a}
\begin{table}[!h]
\centering
\caption{$p$-values $\mathbf{Decimate}$ versus original (not downsampled) signals}
\label{tab:decimate}
\begin{tabular}{cccccc}
\toprule
Factor & Balanced Accuracy & ALS F1-score & Control F1-score & Myopathic F1-score & Significant Degradation \\
\midrule
2 & 1.0000 & 1.0000 & 1.0000 & 1.0000 & No \\
5 & 1.0000 & 1.0000 & 1.0000 & 1.0000 & No \\
10 & 1.0000 & 1.0000 & 1.0000 & 0.9774 & No \\
15 & 1.0000 & 1.0000 & 1.0000 & 1.0000 & No \\
20 & 1.0000 & 1.0000 & 1.0000 & 1.0000 & No \\
25 & 0.7556 & 0.9635 & 0.9086 & 0.5995 & No \\
30 & 0.2622 & 0.9243 & 0.5330 & 0.1070 & No \\
\cellcolor[HTML]{FDEAEA}35 & \cellcolor[HTML]{FDEAEA}0.2444 & \cellcolor[HTML]{FDEAEA}0.9062 & \cellcolor[HTML]{FDEAEA}0.5548 & \bfseries \cellcolor[HTML]{FDEAEA}0.0365 & \cellcolor[HTML]{FDEAEA}Yes \\
40 & 0.1147 & 0.9465 & 0.4452 & \bfseries 0.0128 & Yes \\
45 & 0.1262 & 0.8961 & 0.5110 & \bfseries 0.0128 & Yes \\
50 & 0.0844 & 0.9156 & 0.4235 & \bfseries 0.0043 & Yes \\
55 & \bfseries 0.0240 & 0.8616 & 0.3193 & \bfseries 0.0013 & Yes \\
60 & \bfseries 0.0240 & 0.3808 & 0.0938 & \bfseries 0.0017 & Yes \\
65 & \bfseries 0.0277 & 0.4670 & 0.2106 & \bfseries 0.0021 & Yes \\
70 & \bfseries 0.0177 & 0.2271 & 0.0938 & \bfseries 0.0017 & Yes \\
75 & \bfseries 0.0151 & 0.1514 & 0.0844 & \bfseries 0.0017 & Yes \\
80 & \bfseries 0.0075 & \bfseries 0.0177 & 0.0677 & \bfseries 0.0042 & Yes \\
85 & \bfseries 0.0034 & \bfseries 0.0043 & \bfseries 0.0277 & \bfseries 0.0032 & Yes \\
\bottomrule
\end{tabular}
\end{table}

\begin{table}[!h]
\centering
\caption{$p$-values $\mathbf{LTTB}$ versus original (not downsampled) signals}
\label{tab:lttb}
\begin{tabular}{cccccc}
\toprule
Factor & Balanced Accuracy & ALS F1-score & Control F1-score & Myopathic F1-score & Significant Degradation \\
\midrule
2 & 1.0000 & 1.0000 & 1.0000 & 1.0000 & No \\
5 & 1.0000 & 1.0000 & 1.0000 & 0.8992 & No \\
10 & 1.0000 & 1.0000 & 1.0000 & 1.0000 & No \\
15 & 1.0000 & 1.0000 & 1.0000 & 1.0000 & No \\
20 & 0.7570 & 0.4691 & 1.0000 & 1.0000 & No \\
25 & 0.0958 & 0.0883 & 0.9086 & 0.4155 & No \\
\cellcolor[HTML]{FDEAEA}30 & \bfseries \cellcolor[HTML]{FDEAEA}0.0101 & \bfseries \cellcolor[HTML]{FDEAEA}0.0431 & \cellcolor[HTML]{FDEAEA}0.4221 & \cellcolor[HTML]{FDEAEA}0.0530 & \cellcolor[HTML]{FDEAEA}Yes \\
35 & \bfseries 0.0004 & \bfseries 0.0002 & \bfseries 0.0361 & \bfseries 0.0052 & Yes \\
\bottomrule
\end{tabular}
\end{table}

\begin{table}[!h]
\centering
\caption{$p$-values $\mathbf{MinMaxLTTB}$ versus original (not downsampled) signals}
\label{tab:minmaxlttb}
\begin{tabular}{cccccc}
\toprule
Factor & Balanced Accuracy & ALS F1-score & Control F1-score & Myopathic F1-score & Significant Degradation \\
\midrule
2 & 1.0000 & 1.0000 & 1.0000 & 1.0000 & No \\
5 & 1.0000 & 1.0000 & 1.0000 & 0.4221 & No \\
10 & 1.0000 & 1.0000 & 1.0000 & 1.0000 & No \\
15 & 1.0000 & 1.0000 & 1.0000 & 1.0000 & No \\
20 & 1.0000 & 1.0000 & 1.0000 & 1.0000 & No \\
25 & 0.2031 & 0.2814 & 0.9086 & 0.4542 & No \\
\cellcolor[HTML]{FDEAEA}30 & \bfseries \cellcolor[HTML]{FDEAEA}0.0021 & \bfseries \cellcolor[HTML]{FDEAEA}0.0021 & \cellcolor[HTML]{FDEAEA}0.4221 & \bfseries \cellcolor[HTML]{FDEAEA}0.0167 & \cellcolor[HTML]{FDEAEA}Yes \\
35 & \bfseries 0.0003 & \bfseries 0.0006 & \bfseries 0.0066 & \bfseries 0.0008 & Yes \\
\bottomrule
\end{tabular}
\end{table}

\begin{table}[!h]
\centering
\caption{$p$-values $\mathbf{MinMax}$ versus original (not downsampled) signals}
\label{tab:minmax}
\begin{tabular}{cccccc}
\toprule
Factor & Balanced Accuracy & ALS F1-score & Control F1-score & Myopathic F1-score & Significant Degradation \\
\midrule
2 & 1.0000 & 1.0000 & 1.0000 & 1.0000 & No \\
5 & 1.0000 & 1.0000 & 1.0000 & 0.4221 & No \\
10 & 1.0000 & 1.0000 & 1.0000 & 1.0000 & No \\
15 & 1.0000 & 1.0000 & 1.0000 & 1.0000 & No \\
20 & 1.0000 & 0.9086 & 1.0000 & 1.0000 & No \\
25 & 0.3028 & 0.2814 & 0.9086 & 0.5995 & No \\
\cellcolor[HTML]{FDEAEA}30 & \bfseries \cellcolor[HTML]{FDEAEA}0.0256 & \bfseries \cellcolor[HTML]{FDEAEA}0.0431 & \cellcolor[HTML]{FDEAEA}0.4221 & \cellcolor[HTML]{FDEAEA}0.1070 & \cellcolor[HTML]{FDEAEA}Yes \\
35 & \bfseries 0.0187 & 0.0619 & 0.1419 & \bfseries 0.0181 & Yes \\
40 & \bfseries 0.0067 & 0.0730 & 0.1688 & \bfseries 0.0084 & Yes \\
45 & \bfseries 0.0006 & \bfseries 0.0023 & 0.0956 & \bfseries 0.0043 & Yes \\
50 & \bfseries 0.0009 & \bfseries 0.0013 & \bfseries 0.0384 & \bfseries 0.0043 & Yes \\
\bottomrule
\end{tabular}
\end{table}

\begin{table}[!h]
\centering
\caption{$p$-values $\mathbf{M4}$ versus original (not downsampled) signals}
\label{tab:m4}
\begin{tabular}{cccccc}
\toprule
Factor & Balanced Accuracy & ALS F1-score & Control F1-score & Myopathic F1-score & Significant Degradation \\
\midrule
2 & 1.0000 & 1.0000 & 1.0000 & 1.0000 & No \\
5 & 1.0000 & 1.0000 & 1.0000 & 1.0000 & No \\
10 & 0.9735 & 1.0000 & 1.0000 & 0.2081 & No \\
15 & 1.0000 & 1.0000 & 1.0000 & 0.7570 & No \\
20 & 1.0000 & 1.0000 & 1.0000 & 0.0883 & No \\
\cellcolor[HTML]{FDEAEA}25 & \bfseries \cellcolor[HTML]{FDEAEA}0.0134 & \cellcolor[HTML]{FDEAEA}0.2411 & \cellcolor[HTML]{FDEAEA}0.2675 & \bfseries \cellcolor[HTML]{FDEAEA}0.0029 & \cellcolor[HTML]{FDEAEA}Yes \\
30 & \bfseries 0.0052 & 0.0833 & 0.5245 & \bfseries 0.0134 & Yes \\
35 & \bfseries 0.0215 & 0.1876 & 0.4212 & \bfseries 0.0081 & Yes \\
40 & \bfseries 0.0002 & \bfseries 0.0323 & 0.0619 & \bfseries 0.0002 & Yes \\
45 & \bfseries 0.0002 & 0.0833 & 0.1205 & \bfseries 0.0004 & Yes \\
50 & \bfseries 0.0002 & \bfseries 0.0479 & 0.0554 & \bfseries 0.0002 & Yes \\
55 & \bfseries 0.0002 & \bfseries 0.0013 & \bfseries 0.0081 & \bfseries 0.0003 & Yes \\
\bottomrule
\end{tabular}
\end{table}

\begin{figure*}[t]
    \centering

    \subfloat[ALS F1-score\label{fig:cd_als}]{
        \includegraphics[width=0.45\textwidth]{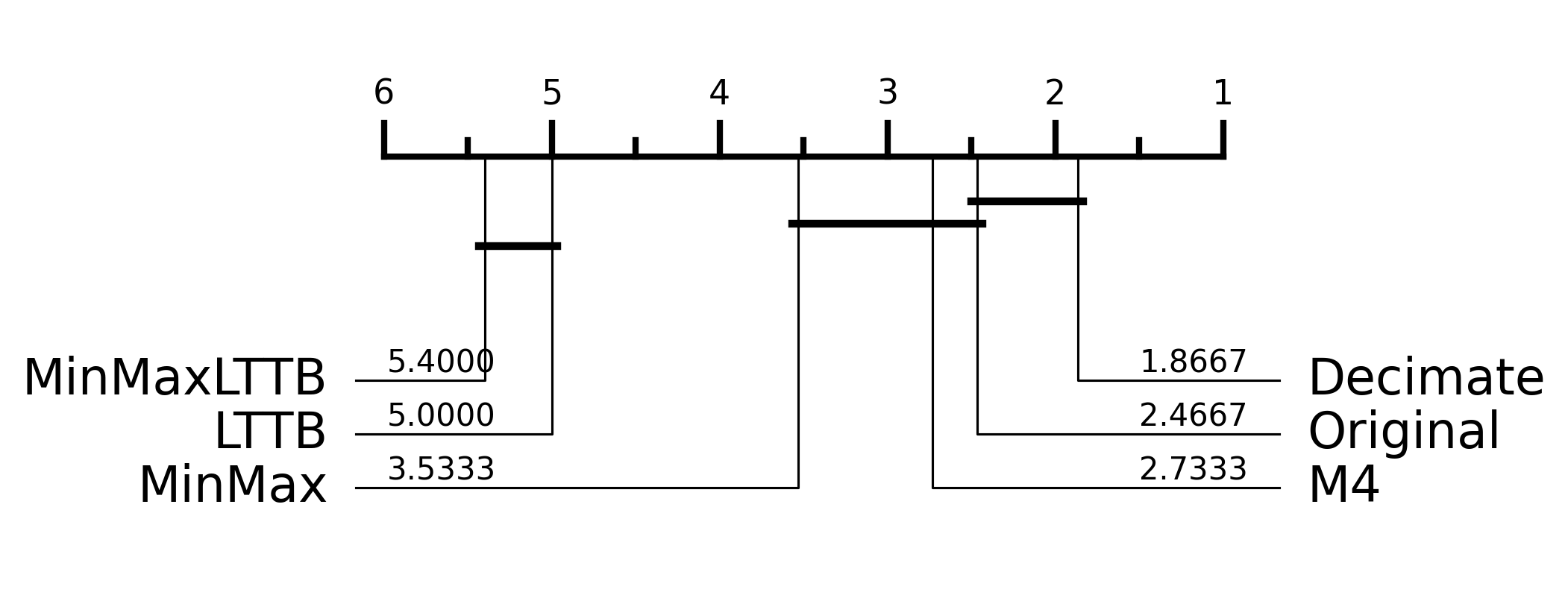}
    }\hspace{-2mm}
    \subfloat[Control F1-score\label{fig:cd_control}]{
        \includegraphics[width=0.45\textwidth]{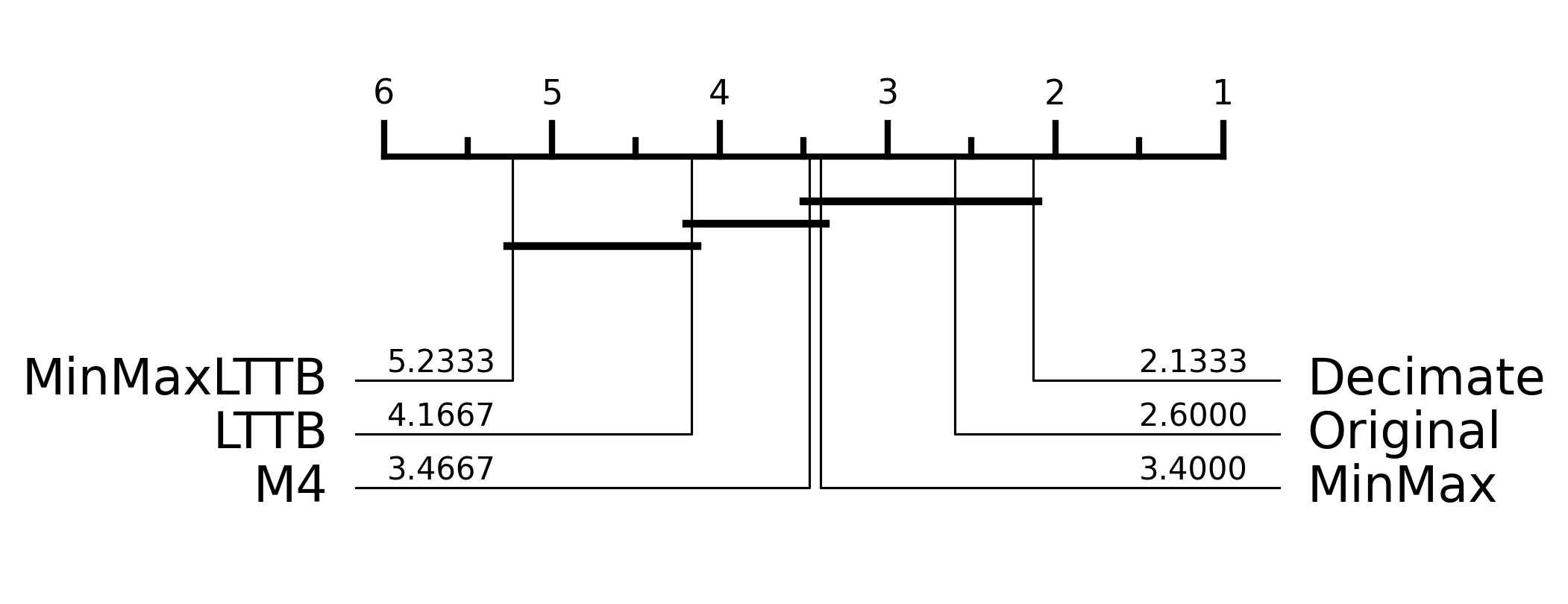}
    }\hspace{-2mm}
    \subfloat[Myopathic F1-score\label{fig:cd_myo}]{
        \includegraphics[width=0.45\textwidth]{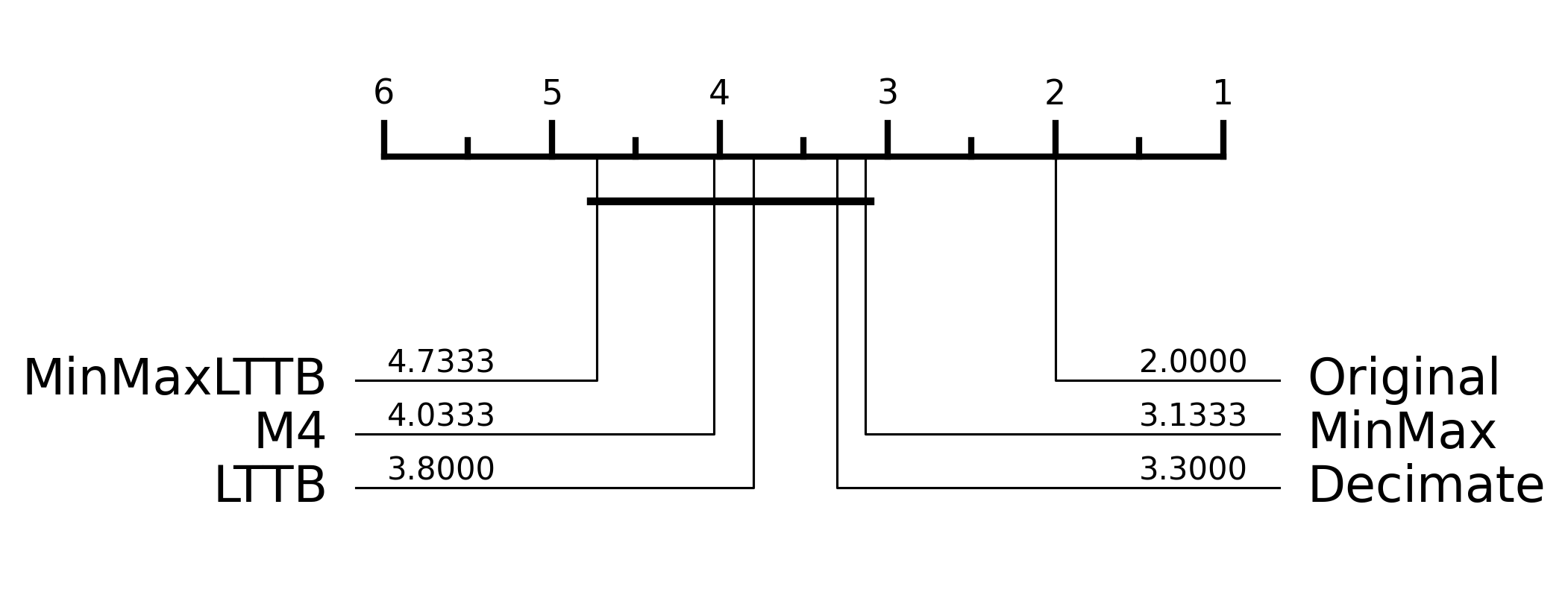}
    }

    \caption{
        Critical difference diagrams comparing downsampling strategies for
        ALS, Control, and Myopathic F1-scores at factor 35.
        Methods connected by a horizontal bar are not significantly different
        under the Wilcoxon signed-rank test with Holm’s sequential correction ($\alpha=0.05$).
    }
    \label{fig:critical_difference}
\end{figure*}

\clearpage
\section{Ranking model performance}
\label{appendix_b}

\begin{figure}[H]
    \centering
    \includegraphics[width=0.8\linewidth]{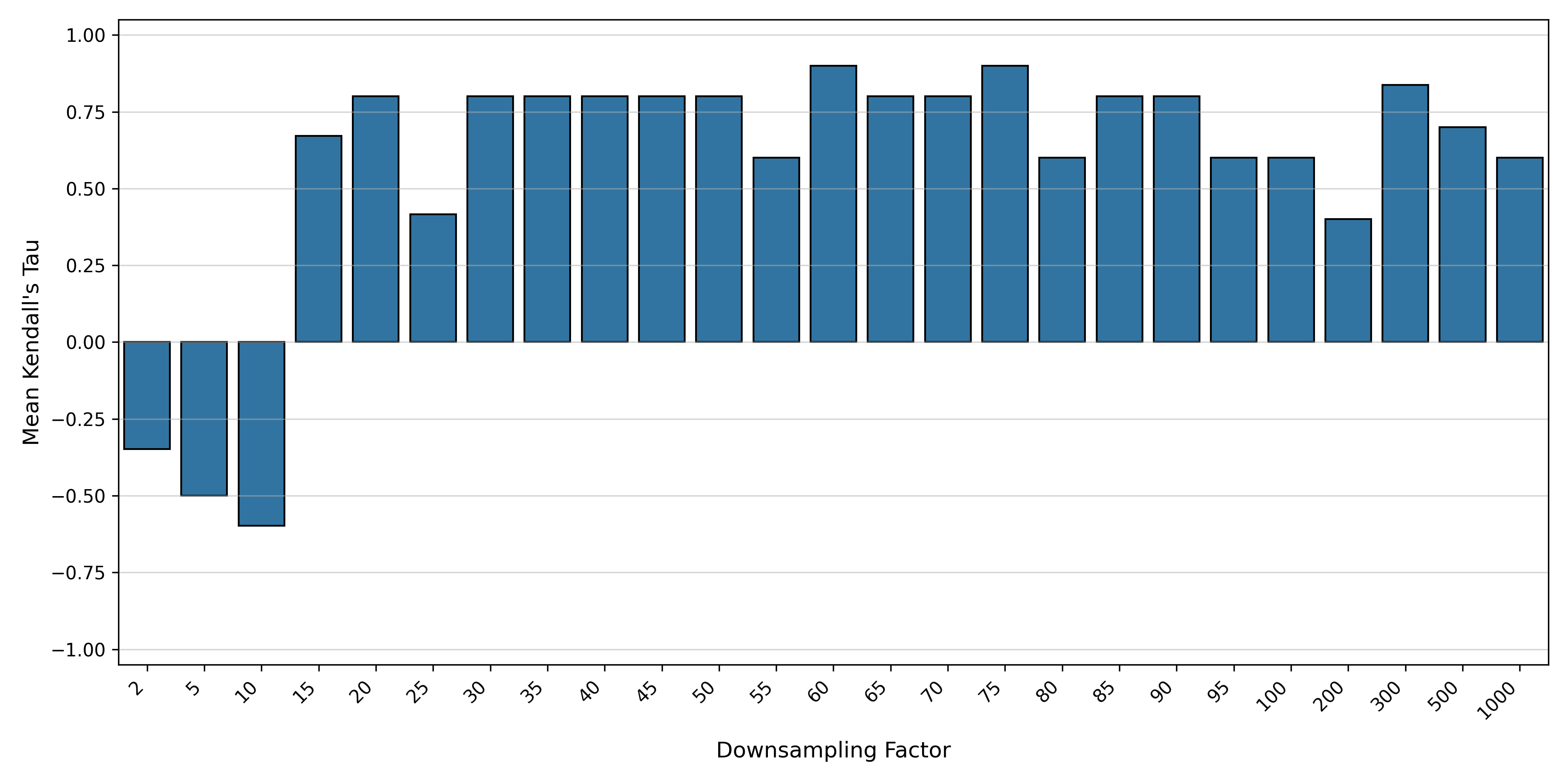}
    \caption{Kendall's $\tau$ correlation between the predicted ranks (using solely distance metrics) and actual ranks (based on the number of wins in balanced accuracy for each possible pairwise combination) between downsampling algorithms at each downsampling factor.}
    \label{fig:ranking_corr}
\end{figure}

\end{document}